\documentclass{article}
\usepackage{amsmath,epsfig}
\usepackage[preprint]{spconfa4}
\usepackage{xspace}
\usepackage{times}
\usepackage{graphics}
\usepackage{amssymb}
\usepackage{pifont}
\usepackage{multirow}
\usepackage{subcaption}
\usepackage{xcolor}
\usepackage[colorlinks,citecolor=green,bookmarks=false,hypertexnames=true]{hyperref} 

\newcommand{\xmark}{\ding{55}}
\newcommand{\cmark}{\ding{52}}
\newcommand{\wa}{.115}
\newcommand{\wb}{.75}
\newcommand{\waqmesh}{.15}
\newcommand{\wbqmesh}{1.0}
\newcommand{\waqup}{.15}
\newcommand{\wbqup}{1.1}
\newcommand{\wqreal}{0.23}
\newcommand{\hqreal}{3.5cm}
\newcommand{\tablesize}{0.62}
\newcommand{\wden}{0.11}
\newcommand{\heden}{1.7cm}
\newcommand{\wnoi}{0.14}
\newcommand{\henoi}{1.75cm}
\newcommand{\wfail}{0.2}

\newcommand*{\eg}{$e.g.$}
\newcommand*{\ie}{$i.e.$}

\newcommand{\hefail}{3.8cm}
\newcommand{\hefailb}{3.2cm}

\let\OLDthebibliography\thebibliography
\renewcommand\thebibliography[1]{
  \OLDthebibliography{#1}
  \setlength{\parskip}{0pt}
  \setlength{\itemsep}{0pt plus 0.3ex}
}

\begin{document}\sloppy

\def\x{{\mathbf x}}
\def\L{{\cal L}}

\title{``Zero-Shot'' Point Cloud Upsampling}
\name{Kaiyue Zhou$^{1,2}$\thanks{$\ast$ Corresponding Author. The source code is available at \url{https://github.com/ky-zhou/ZSPU}.}, Ming Dong$^{1 \ast}$, and Suzan Arslanturk$^{1}$}
\address{$^{1}$Department of Computer Science, Wayne State University, USA\\ $^{2}$Department of Electronic Engineering, Tsinghua University, China\\ \{kyzhou, mdong, suzan.arslanturk\} @wayne.edu}

\maketitle

\begin{abstract}
Recent supervised point cloud upsampling methods are restricted by the size of training data and are limited in terms of covering all object shapes. Besides the challenges faced due to data acquisition, the networks also struggle to generalize on unseen records. In this paper, we present an internal point cloud upsampling approach at a holistic level referred to as ``Zero-Shot'' Point Cloud Upsampling (ZSPU). Our approach is data agnostic and relies solely on the internal information provided by a particular point cloud without patching in both self-training and testing phases. This single-stream design significantly reduces the training time by learning the relation between low resolution (LR) point clouds and their high (original) resolution (HR) counterparts. This association will then provide super resolution (SR) outputs when original point clouds are loaded as input. ZSPU achieves competitive/superior quantitative and qualitative performances on benchmark datasets when compared with other upsampling methods.
\end{abstract}
\begin{keywords}
Internal, holistic, zero-shot, upsampling
\end{keywords}
\section{Introduction}
\begin{figure}[t]
\centering
\begin{subfigure}{\wa\textwidth}
  \centering
  \includegraphics[width=\wb\linewidth]{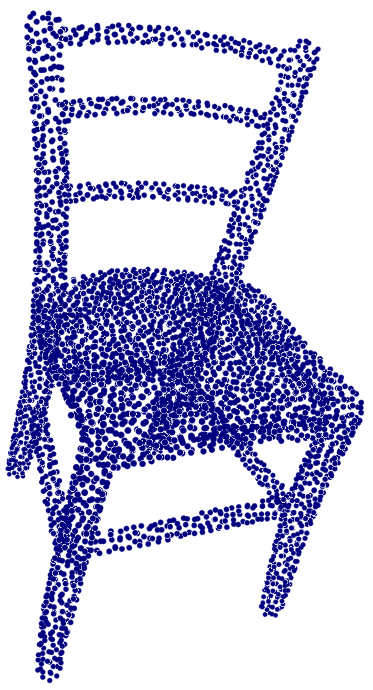}  
  \label{fig1:input1}
\end{subfigure}
\begin{subfigure}{\wa\textwidth}
  \centering
  \includegraphics[width=\wb\linewidth]{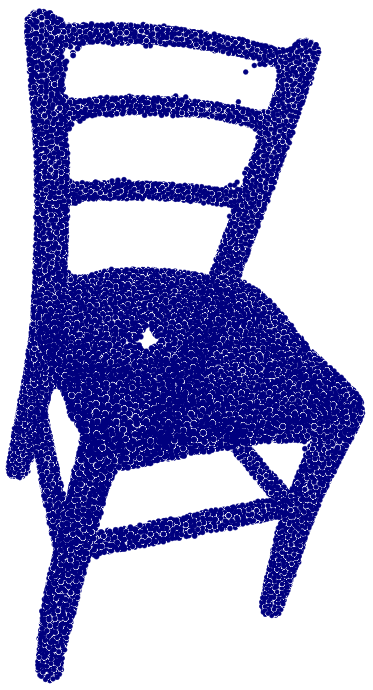}  
  \label{fig1:pugan1}
\end{subfigure}
\begin{subfigure}{\wa\textwidth}
  \centering
  \includegraphics[width=\wb\linewidth]{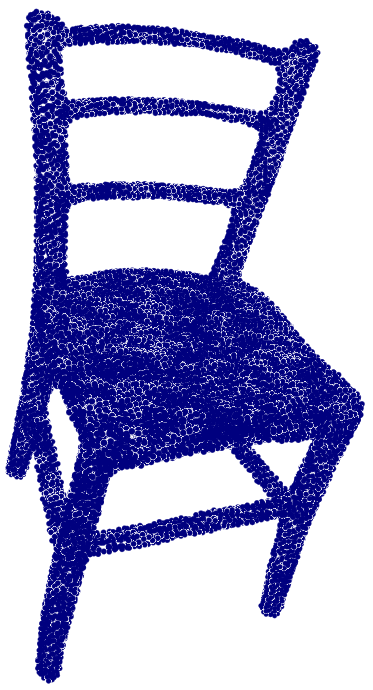}  
  \label{fig1:ZSPU1}
\end{subfigure}
\begin{subfigure}{\wa\textwidth}
  \centering
  \includegraphics[width=\wb\linewidth]{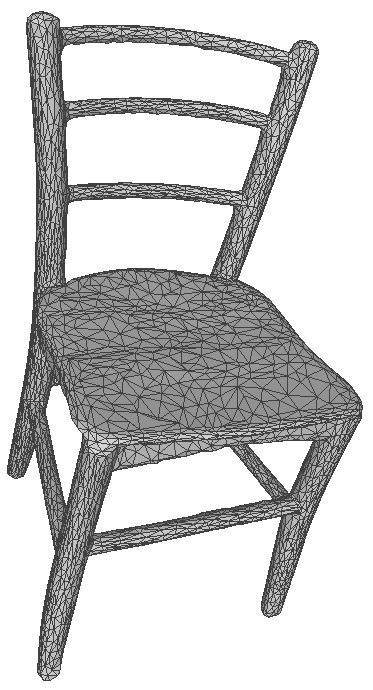}  
  \label{fig1:mesh1}
\end{subfigure}

\begin{subfigure}{\wa\textwidth}
  \centering
  \includegraphics[width=\wb\linewidth]{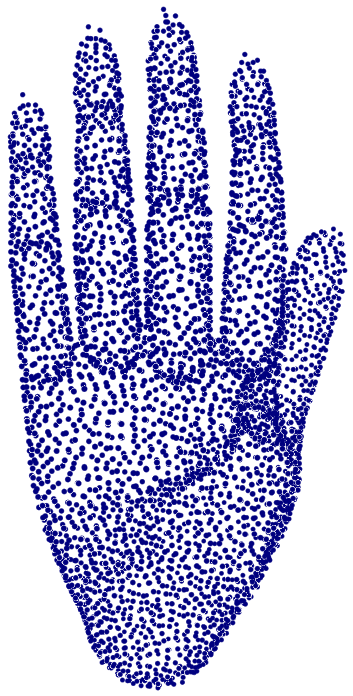}
  \caption{Input}
  \label{fig1:input2}
\end{subfigure}
\begin{subfigure}{\wa\textwidth}
  \centering
  \includegraphics[width=\wb\linewidth]{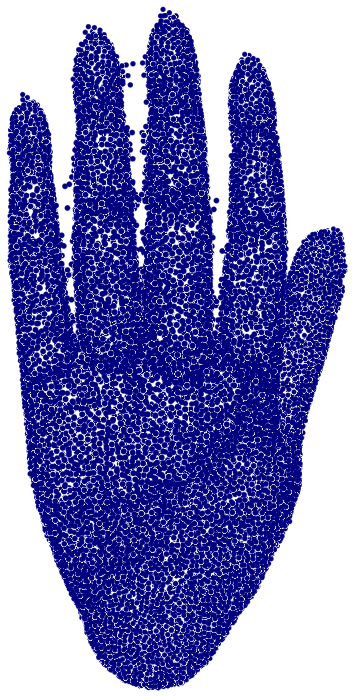}
  \caption{PU-GAN}
  \label{fig1:pugan2}
\end{subfigure}
\begin{subfigure}{\wa\textwidth}
  \centering
  \includegraphics[width=\wb\linewidth]{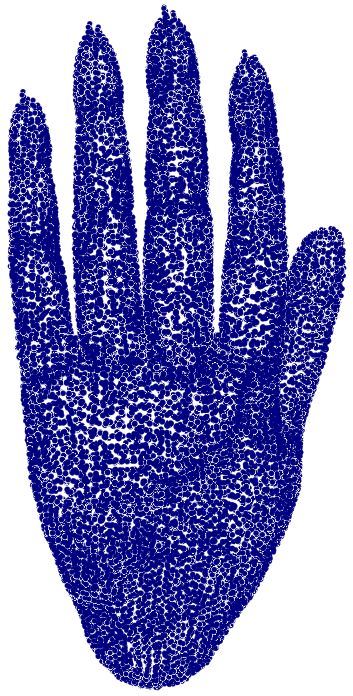}
  \caption{ZSPU}
  \label{fig1:ZSPU2}
\end{subfigure}
\begin{subfigure}{\wa\textwidth}
  \centering
  \includegraphics[width=\wb\linewidth]{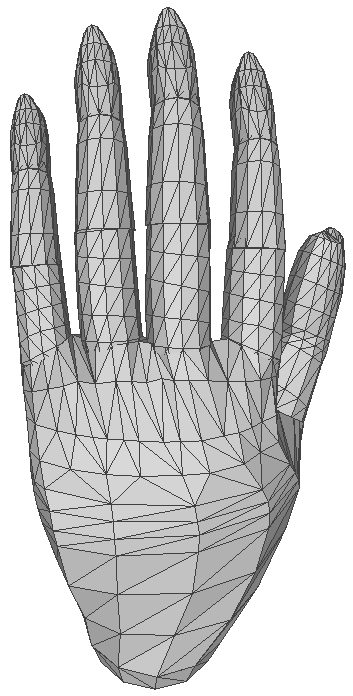}  
  \caption{Mesh}
  \label{fig1:mesh2}
\end{subfigure}
\caption{(a) Input point cloud. (b) Patch based method sacrifices partial reconstruction capacity to generate more uniform distribution of points, while (c) our method tends to recover the original shape. (d) Ground truth mesh.}
\label{fig:fig1}
\end{figure}

In 3D computer vision domain, point cloud upsampling has recently achieved outstanding performance through supervised deep learning methods~\cite{yu2018pu, yu2018ec, yifan2019patch, li2019pugan}. However, these methods require tremendous amounts of training data to understand various structural shapes. While current 3D datasets for upsampling tasks from PU-Net~\cite{yu2018pu}, 3PU~\cite{yifan2019patch}, PU-GAN~\cite{li2019pugan}, Dis-PU~\cite{li2021point}, and PU-GCN~\cite{qian2021pu} covered a rich variety of objects, the absence of certain structures observed in real world scenarios (\eg, trees, vehicles, animals, etc.) would limit the learning process. Particularly, although 3PU, PU-GAN, Dis-PU, and PU-GCN have proposed a patch-based scheme to improve the local uniformity of points for both training and testing phases, there still exists two potential limitations that restrict them from being applied to all sizes and structures of shapes. First, the parameters such as the number of points in each patch (NoPP) and the number of patches (NoP) need to be carefully set for various shapes to avoid producing problematic point clouds, (\eg, in Fig.~\ref{fig1:pugan2} (top), here the patches generated by farthest-point-sampling (FPS) algorithm are not sufficient to preserve the entire shape). Second, the nearby curvature might be disturbed by points in a patch with high geodesic distances \eg, fingers, as shown in Fig.~\ref{fig1:pugan2} (bottom). Finally, pre-processing such patches is not trivial, which requires to extract 3D coordinates (sometimes even the geodesic~\cite{he2019geonet} or edge~\cite{yu2018ec} information) from the existing mesh data.

In this paper, we present a ``zero-shot'' point cloud upsampling (ZSPU) framework to internally learn the self-aware representation of a holistic point cloud. This design skips the setup of parameters like NoPP and NoP to avoid potentially fallacious topological mappings and effectively preserves the shape of the target point cloud. We summarize our \textbf{contributions} as follows: (i) To the best of our knowledge, ZSPU is the first holistic and internal point cloud upsampling method. Our holistic design adapts the target shape complexity automatically and avoids the intricate settings and patch pre-processing in patch-based methods. Additionally, the internal learning only requires one input point cloud, which significantly reduces the amount of training samples and time required by other supervised methods. (ii) Quantitatively, we achieve competitive performances on a benchmark dataset when compared with existing point cloud upsampling methods, and our method outperforms those methods on another benchmark dataset with unseen categories. (iii) Qualitatively, ZSPU preserves better local detail and curvature without losing much uniformity performance when compared with patch-based upsampling methods and thus is well suited to handle complex scenes obtained from real-world scans.
%
%

\section{Related Work}
In this section, we briefly review the methods of point cloud upsampling and single image super resolution.

\subsection{Point Cloud Upsampling}
PointNet~\cite{qi2017pointnet} and PointNet++~\cite{qi2017pointnet++} have motivated multiple point cloud networks targeting upsampling in the past few years. Taking advantage of multi-scale grouping (MSG) and feature interpolation, PU-Net~\cite{yu2018pu} first constructed a deep learning network that hierarchically upsamples the point clouds. Thereafter, EC-Net~\cite{yu2018ec} took edge and surface information as prior knowledge, such that the upsampled point cloud forces the points to consolidate along the edges and surfaces. By leveraging geodesic distance as an additional feature, GeoNet~\cite{he2019geonet} learned surface topology around each point. 3PU~\cite{yifan2019patch} first introduced the progressive trend with skip connections that constructs features from previous layers. Dis-PU~\cite{li2021point} split the upsampling framework to a coarse point generator and a refiner that regressed offset to the points. AR-GCN~\cite{wu2019point}, PU-GAN~\cite{li2019pugan}, and PU-GCN~\cite{qian2021pu} applied adversarial learning to point cloud upsampling. PU-GCN designed a novel NodeShuffle module to rearrange the points and their neighbor features.

In general, the aforementioned external learning methods perform well to shapes with smooth surfaces or small curvatures. However, they are all patch-based approaches and suffer from the potential mismatch between the patching scheme and the target shape, especially for complex shapes with high curvatures. Additionally, they may require different hyperparameters (\eg, NoPP and NoP) for different shapes.

\subsection{Single Image Super Resolution}
External learning is the most common manner in supervised learning trained with massive data. In contrast, internal learning has a pivotal role on problems that are restricted to relatively smaller data size.

Single image super resolution (SISR) aims to reconstruct an HR image from a degraded LR query. Most recent studies adopted internal examples~\cite{ma2020structure, niu2020single} for this task. Others utilized external information including zoomed camera image pairs~\cite{zhang2019zoom} and texture transfer learning~\cite{zhang2019image} to gain the extra knowledge for the network. A common characteristic among these studies is that they are not focusing on generating higher (super) resolution than the ground truth or the HR image.

``Zero-shot'' super resolution (ZSSR)~\cite{shocher2018zero} was initially proposed to perform super resolution (SR) on a single image. This approach synchronously trains and tests the network with augmented pairs of LR-HR images extracted from the test image such that the synthesized image has a higher resolution than the HR image. By exploiting the internal recurrence of information within a single image and training an image-specific model at test time, ZSSR achieved competitive/superior results when compared with the external learning methods. Inspired by ZSSR, we use the term ``zero-shot'' in this paper to indicate holistic and internal upsampling of point clouds.

\section{Method}
\begin{figure}[ht]
\begin{center}
    \includegraphics[width=1.0\linewidth]{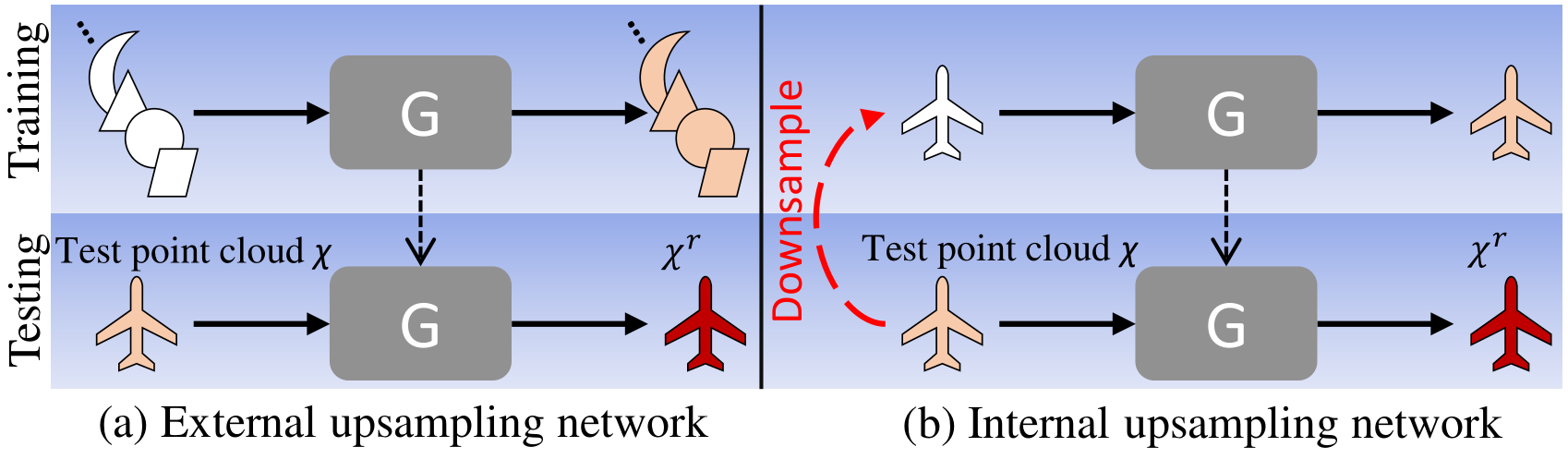}
\end{center}
    \caption{``Zero-Shot'' point cloud upsampling. (a) External methods are first trained on a large number of patches. The test point cloud (whole or patched) $\chi$ is then fed into the network. (b) ZSPU is trained internally on the test point cloud itself. The resulting point cloud specific network is then applied on the original point cloud to produce a denser output.}
\label{fig:overview}
\end{figure}

Given a point set $\chi=\{x_i\}$, where $x_i \in \mathbb{R}^d$ and $i=1,2,...,N$, we aim at generating a denser point set $\chi^r=\{x_j\}$, where r is the upsampling ratio and $j=1,2,...,rN$. Here, we use $d=3$, \ie, xyz coordinate in Euclidean space. As the only available training data is $\chi$, we apply either a random (non-ideal) or FPS (ideal) kernel to downsample $\chi$ to several LR versions, \ie, $\chi^{{LR}_i}, i=1,2,...,B$, where $B$ is configured by the user during self-training. ZSPU provides an internal, holistic learning framework for point cloud upsampling. The key difference between ZSPU and existing external, patch-based methods is illustrated in Fig.~\ref{fig:overview}, where a darker object color represents a higher resolution.

\subsection{Architecture}
ZSPU is a generative adversarial network (GAN) conditioning on a specific point cloud. More details of the architecture are provided in supplementary materials.
\begin{figure}[ht]
\begin{center}
    \includegraphics[width=1.0\linewidth]{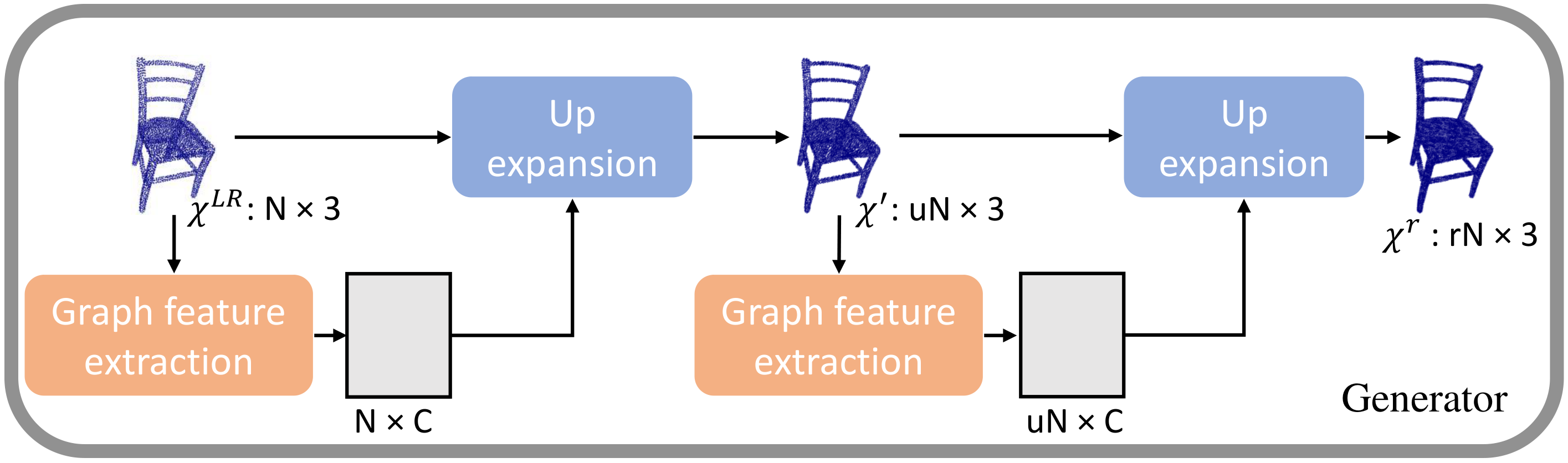}
\end{center}
    \caption{The input point cloud has size of $N$. After progressive up expansions, the output $\chi^r$ has $rN$ points. $u$ and $r$ are the progressive and overall upsampling ratios.}
\label{fig:4:arch}
\end{figure}

\textbf{Generator.} As shown in Fig.~\ref{fig:4:arch}, the generator takes LR children ${\chi^{{LR}_i}}$ as input, and progressively upsamples with ratio 2 until the upsampling ratio is reached. This procedure repeats $t$ times, where $t={log}(r)$. This progressive learning approach supports changing the overall upsampling ratio $r$ without modifying the network. We use $\chi^r$ to denote the output (and hereafter).

\textbf{Discriminator.} We adopt the module in \cite{li2019pugan} to distinguish the generated $\chi^r$ from the real target $\chi^{gt}$. The global and point-wise features are first concatenated for better understanding the overall shape and details of the object. An attention module thereafter captures the point-wise similarities given the flattened features. Finally, a fidelity score is predicted via an aggregation of multilayer perceptron, max pooling, and dense layers.

\subsection{Loss Function}
Our joint loss function of the generator can be written as:
\begin{equation}\label{eq:compound}
L = \alpha L_{G}+\beta L_{EMD}(\chi^r, \chi^{gt})+\upsilon L_{uni} + \lambda L_{rep}(\chi^r)+\omega {||\theta ||}^2,
\end{equation}
in where, $||\theta ||$ is the regularization term, and $\alpha$, $\beta$, $\upsilon$, $\lambda$, and $\omega$ are the weights, $L_{G}$, $L_{EMD}$, $L_{uni}$, and $L_{rep}$ are the adversarial, reconstruction, uniform, and repulsion losses, respectively. The details of loss functions are described in the supplementary material.

\subsection{Holistic vs. Patching}
\begin{figure}[ht]
\begin{center}
    \includegraphics[width=1.0\linewidth]{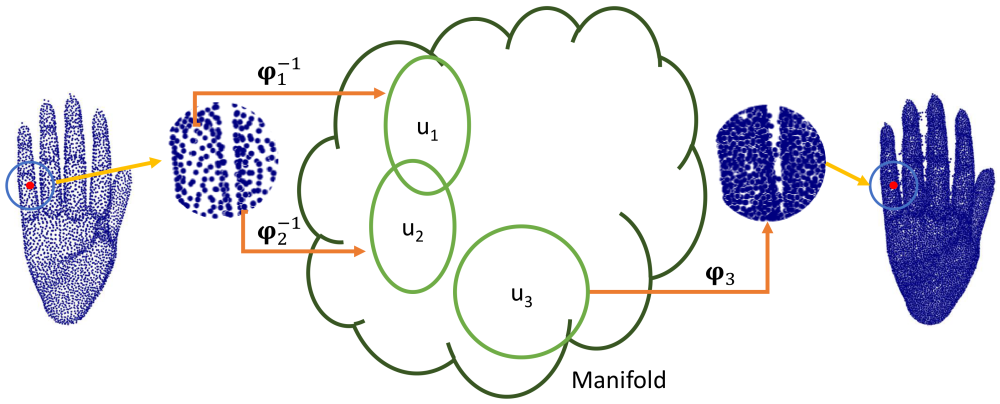}
\end{center}
    \caption{In this example, the little finger and the ring finger are considered as a single shape since the manifold fails to map the two parts ($u_1$ and $u_2$) simultaneously. Instead, $u_3$ is selected as the corresponding subset in topology $T_M$, such that the mixture is occurred by $\varphi_{3}(u_3)$.}
\label{fig:manifold}
\end{figure}
Unlike patch-based methods (\ie, subdividing a point cloud into patches during testing and then merging upsampled patches later using FPS), our approach tends to learn and predict in a holistic style. In patch-based methods, the motivation is to map the testing patches to the similar pairs in the training set. However, the formation of patches restricts the accuracy of reconstruction of the entire shape, \ie, a set formed by kNN of a centroid might have extremely large geodesic distance while having a small euclidean distance. This requires patch based methods to select parameters (\eg, NoPP and NoP) carefully. In the case of a patch containing points from different parts of the object (\eg, two fingers), the network might fail seeking for the correct subsets in the manifold for these parts.

We define $M$ as manifold, $T_M$ as the topology for $M$, and $\Lambda$ as the transition functions between Euclidean and topological spaces. Specifically, in $\Lambda$, we denote $\varphi^{-1}$ and $\varphi$ as the transition functions into and out of the topological space, respectively. Given $(M, T_M, \Lambda)$, this procedure can be formally written as 
\begin{equation}
u \in T_M: \exists \varphi: u \rightarrow \varphi(u) \subset \mathbb{R}^d.
\end{equation}

In Fig.~\ref{fig:manifold}, we illustrate an example when an input patch $P_{in}$ is mapped into the topological space, its corresponding subsets $\{u_1, u_2\} = \varphi^{-1}(P_{in})$ cannot be accurately located since $\{u_1 + u_2\} \geqslant \{u_1 \cup u_2\}$. Alternatively, the network could select $u_3$ to represent $\{u_1 \cup u_2\}$ but $\varphi(\{u_1 \cup u_2\}) \neq \varphi(u_3)$. Empirically, this phenomenon can be alleviated by using more training samples. However, it is hard to cover all shapes in real-world applications.

In our holistic setting, $P_{in}$ denotes the entire point cloud, such that $u_h = \varphi^{-1}(P_{in})$ can represent the topology $T_M$ of $P_{in}$. The transition $\varphi(u_h)$ to Euclidean space is then directly conducted without any potential mapping problem in the topological space.

\section{Experiments}
\subsection{Datasets}
We use the testing set (20 objects) in PU-GAN~\cite{li2019pugan} and select 20 representative objects in Princeton Shape Benchmark~\cite{shilane2004princeton} (denoted as ${Data}_{PU}$ and ${Data}_{PS}$, respectively) for our quantitative experiments. Specifically, due to ${Data}_{PU}$ being non-uniform, we generate 4,096 points (HR) using the meshes in ${Data}_{PU}$ through Poisson sampling algorithm. If not specified, we apply the non-ideal kernel to downsample 4,096 points (HR) to 1,024 (LR) through a random selection in all experiments. The ideal kernel uses FPS to downsample the point sets, which is denoted by ZSPU-I in our experiments. There are $B$ pairs of LR and HR point clouds, which in our test is set as 12 for the purpose of performance and computational efficiency. In our experiments, the patched training data from other works are not taken into consideration. 

\subsection{Implementation Detail}
To train ZSPU, we use Adam optimizer with learning rate 0.001 for the generator and 0.0001 for the discriminator, and batch size 12 for 50 epochs. Data augmentation is applied to all LR inputs, \ie, rotation, jittering, shifting, and scaling. The weights in Eq.~\ref{eq:compound} are empirically set at 0.005, 1, 0.1, 0.01, and 0.01, respectively. Our network is implemented on TensorFlow 1.15 and trained by NVIDIA Titan RTX GPU.

\subsection{Evaluation Metric}
For the quantitative evaluation, we report the following widely-adopted metrics \cite{li2019pugan}: Chamfer distance (CD), Hausdorff distance (HD), point-to-surface (P2F) distance in average and its standard deviation, and uniformity on different radii in a unit area. To compare distances with ground truth, we uniformly sample 16,384 points from the ground truth mesh using the Poisson sampling algorithm. Lower values in the evaluation metrics represent better performances.

\subsection{Quantitative Comparison}
\begin{figure*}[ht]
\centering
\begin{subfigure}{\waqmesh\textwidth}
  \includegraphics[width=\wbqmesh\linewidth]{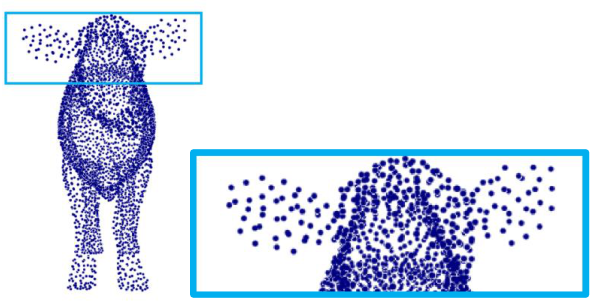}
  \caption{Input}
\end{subfigure}
\hspace*{\fill}
\begin{subfigure}{\waqmesh\textwidth}
  \includegraphics[width=\wbqmesh\linewidth]{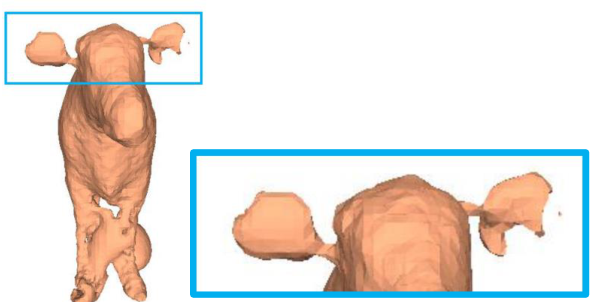}
  \caption{PU-Net}
\end{subfigure}
\hspace*{\fill}
\begin{subfigure}{\waqmesh\textwidth}
  \includegraphics[width=\wbqmesh\linewidth]{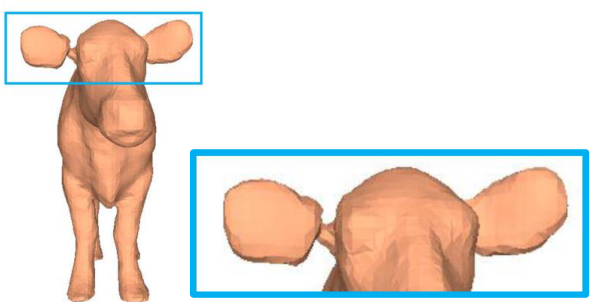}
  \caption{AR-GCN}
\end{subfigure}
\hspace*{\fill}
\begin{subfigure}{\waqmesh\textwidth}
  \includegraphics[width=\wbqmesh\linewidth]{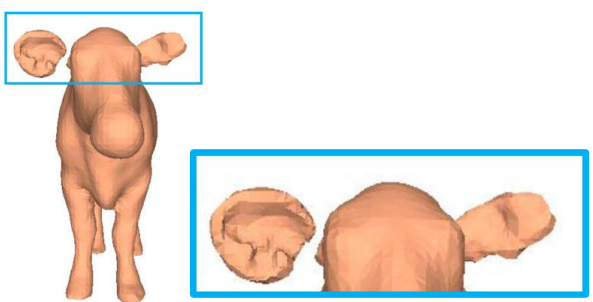}
  \caption{3PU}
\end{subfigure}
\hspace*{\fill}
\begin{subfigure}{\waqmesh\textwidth}
  \includegraphics[width=\wbqmesh\linewidth]{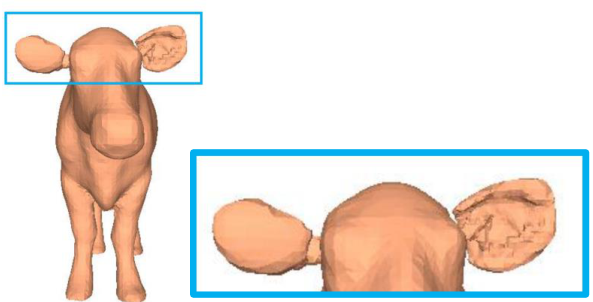}
  \caption{PU-GAN}
\end{subfigure}
\hspace*{\fill}
\begin{subfigure}{\waqmesh\textwidth}
  \includegraphics[width=\wbqmesh\linewidth]{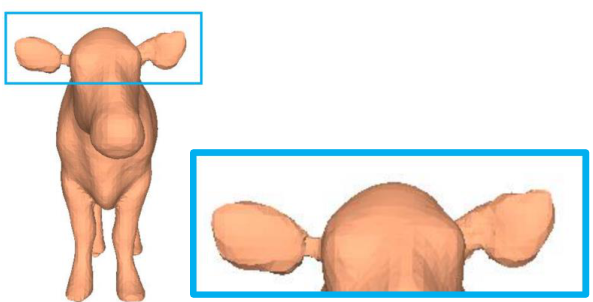}
  \caption{ZSPU}
  \label{figq4:zs}
\end{subfigure}
\caption{Comparison of surface reconstruction results. Our method has better local reconstruction than the patch-based methods even for shapes with smooth surfaces.}
\label{fig:figq}
\end{figure*}
\begin{figure*}[ht]
\centering
\begin{subfigure}{\waqup\textwidth}
  \includegraphics[width=\wbqup\linewidth]{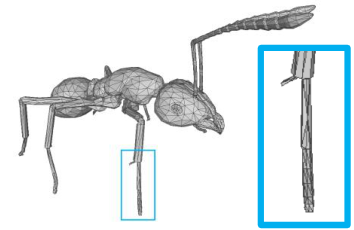}
  \caption{Mesh}
\end{subfigure}
\hspace*{\fill}
\begin{subfigure}{\waqup\textwidth}
  \includegraphics[width=\wbqup\linewidth]{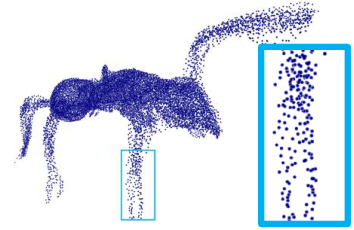}
  \caption{PU-Net}
\end{subfigure}
\hspace*{\fill}
\begin{subfigure}{\waqup\textwidth}
  \includegraphics[width=\wbqup\linewidth]{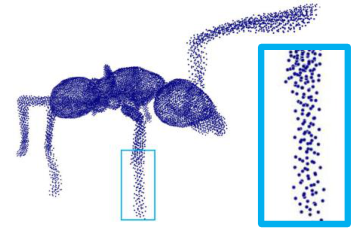}
  \caption{AR-GCN}
\end{subfigure}
\hspace*{\fill}
\begin{subfigure}{\waqup\textwidth}
  \includegraphics[width=\wbqup\linewidth]{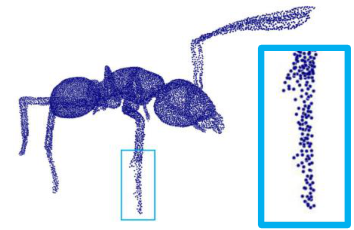}
  \caption{3PU}
\end{subfigure}
\hspace*{\fill}
\begin{subfigure}{\waqup\textwidth}
  \includegraphics[width=\wbqup\linewidth]{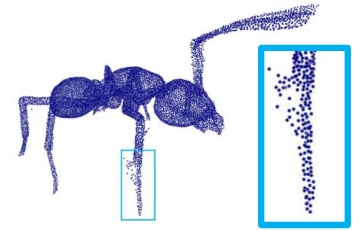}
  \caption{PU-GAN}
\end{subfigure}
\hspace*{\fill}
\begin{subfigure}{\waqup\textwidth}
  \includegraphics[width=\wbqup\linewidth]{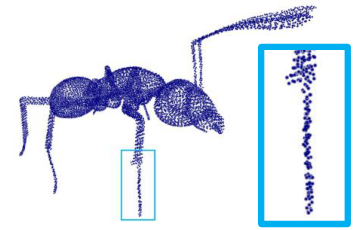}
  \caption{ZSPU}
\end{subfigure}
\caption{Comparison of synthesis for complex shapes. Visually, more local details are preserved by our method.}
\label{fig:figq2}
\end{figure*}
\begin{figure*}[h!]
\centering
\begin{subfigure}{\waqmesh\textwidth}
  \includegraphics[width=\wbqmesh\linewidth]{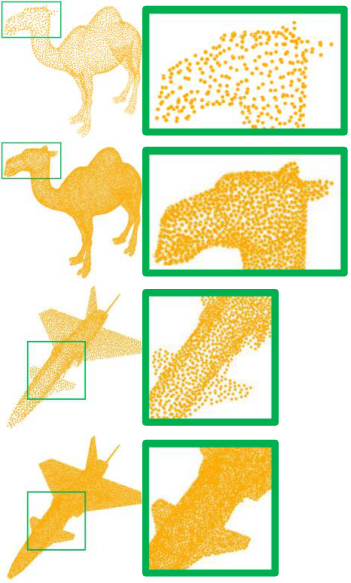}
  \caption{Input \& GT}
\end{subfigure}
\hspace*{\fill}
\begin{subfigure}{\waqmesh\textwidth}
  \includegraphics[width=\wbqmesh\linewidth]{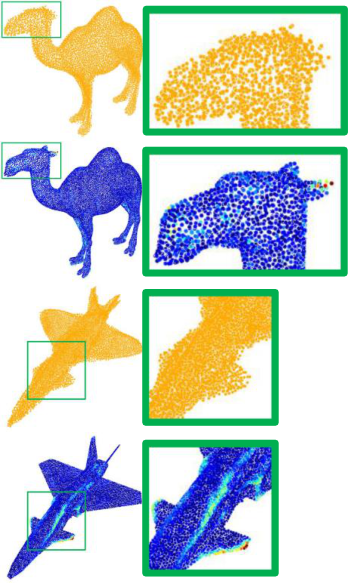}
  \caption{PU-Net}
\end{subfigure}
\hspace*{\fill}
\begin{subfigure}{\waqmesh\textwidth}
  \includegraphics[width=\wbqmesh\linewidth]{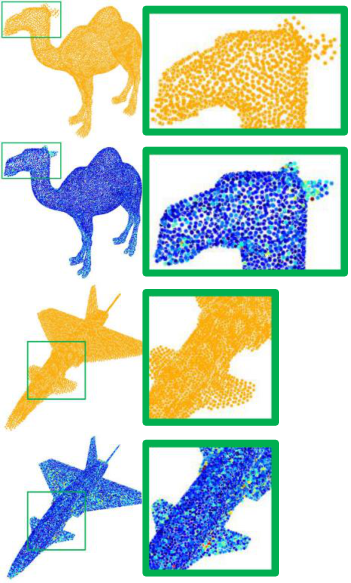}
  \caption{AR-GCN}
\end{subfigure}
\hspace*{\fill}
\begin{subfigure}{\waqmesh\textwidth}
  \includegraphics[width=\wbqmesh\linewidth]{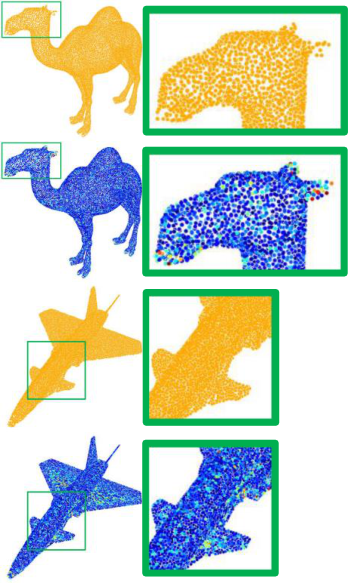}
  \caption{3PU}
\end{subfigure}
\hspace*{\fill}
\begin{subfigure}{\waqmesh\textwidth}
  \includegraphics[width=\wbqmesh\linewidth]{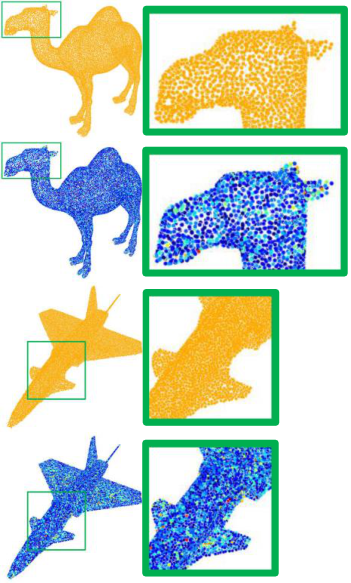}
  \caption{PU-GAN}
\end{subfigure}
\hspace*{\fill}
\begin{subfigure}{\waqmesh\textwidth}
  \includegraphics[width=\wbqmesh\linewidth]{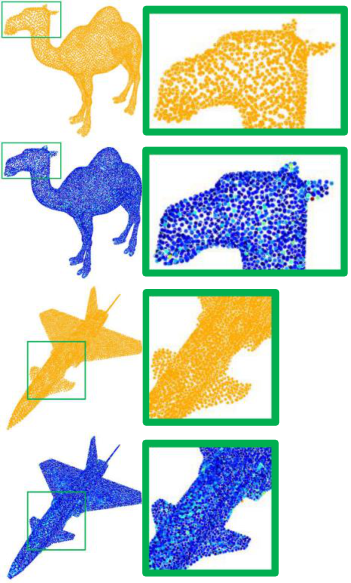}
  \caption{ZSPU}
  \label{figerr:zs}
\end{subfigure}
  \includegraphics[width=0.5\linewidth]{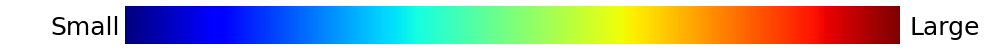}
\caption{Comparison of error maps. The colors reveal the nearest distance from each target point to the generated point. ZSPU tends to preserve more local curvatures while maintaining competitive performance on smooth regions.}
\label{fig:errormap}
\end{figure*}

Although unfair, we quantitatively compare ZSPU with state-of-the-art external learning networks. We use their pre-trained models to test on ${Data}_{PU}$ and ${Data}_{PS}$. Here, we manually set up NoPP and NoP during testing to obtain visually complete point clouds for patch-based methods (\ie, 3PU, PU-GAN, Dis-PU, and PU-GCN) as they significantly affect the reconstruction performance. Empirically, these hyperparameters fluctuate exceedingly with the shape complexity of the object or the number of objects in a target scene, making it nontrivial to apply patch-based methods straightly on all target categories. We highlight the best results with \textcolor{red}{red} color and the second best with \textcolor{blue}{blue} color in all quantitative results.
\begin{table}[t]
\begin{center}
\scalebox{\tablesize}{
\begin{tabular}{c||cc|cc|c|cc|c|c|c}
\hline
\multirow{2}{*}{Methods} & CD & HD & \multicolumn{2}{c|}{P2F ($10^{-3}$)} & Uniformity & Train & Test & \multirow{2}{*}{Epoch} & \multirow{2}{*}{Time} & \# of \\
& \multicolumn{2}{c|}{($10^{-3}$)} & $\mu$ & $\sigma$ & 0.4\% ($10^{-3}$) & \multicolumn{2}{c|}{Holistic} & & & param \\
\hline\hline
PU-Net & 0.38 & 3.67 & 8.19 & 6.65 & 6.36 & \xmark & \textcolor{green}{\cmark} & 120 & 4.5h & 814k \\
AR-GCN & 0.23 & 1.78 & 3.02 & 3.52 & 1.29 & \xmark & \textcolor{green}{\cmark} & 120 & 6.2h & 822k \\
3PU & 0.21 & 1.90 & \textcolor{blue}{1.72} & 2.21 & 1.32 & \xmark & \xmark & 400 & 27h & 304k \\
PU-GAN & \textcolor{red}{0.17} & 1.76 & \textcolor{red}{1.05} & 1.92 & \textcolor{red}{0.55} & \xmark & \xmark & 100 & 25h & 684k \\
Dis-PU & 0.15 & 1.40 & 1.19 & \textcolor{blue}{1.86} & \textcolor{blue}{0.58} & \xmark & \xmark & 400 & 80h & 1047k \\
PU-GCN & 0.26 & 2.62 & 2.15 & 3.01 & 1.75 & \xmark & \xmark & 100 & 9h & 542k \\
\hline
ZSPU & \textcolor{blue}{0.19} & \textcolor{blue}{1.11} & 2.12 & 2.21 & 2.24 & \textcolor{green}{\cmark} & \textcolor{green}{\cmark} & \textcolor{red}{50} & \textcolor{red}{96s} & 310k \\
ZSPU-I & 0.20 & \textcolor{red}{1.10} & 1.89 & \textcolor{red}{1.79} & 1.89 & \textcolor{green}{\cmark} & \textcolor{green}{\cmark} & \textcolor{red}{50} & \textcolor{blue}{98s} & 310k \\
\hline
\end{tabular}}
\end{center}
\caption{Quantitative comparisons with supervised upsampling networks on ${Data}_{PU}$.}
\label{tab:quantpu}
\end{table}

Table~\ref{tab:quantpu} shows the quantitative comparisons on ${Data}_{PU}$. The epoch and time are reported by the corresponding articles without retraining on our end. In terms of HD, our ZSPU outperforms all other supervised methods. Results show that our method has competitive performance with supervised methods regarding CD, P2F, and uniformity metrics. Note that our training time is significantly less than the others, and our method avoids selecting hyper-parameters (\ie, NoPP and NoP) for miscellaneous scenarios. Also in Table~\ref{tab:quantpu}, we report the number of trainable parameters for the upsampling methods with 4096 points as the input. ZSPU generally has a smaller size than other models. 
\begin{table}[t]
\begin{center}
\scalebox{\tablesize}{
\begin{tabular}{c||cc|cc|ccccc}
\hline
\multirow{2}{*}{Methods} & CD & HD & \multicolumn{2}{c|}{P2F ($10^{-3}$)} & \multicolumn{5}{c}{Uniformity for different p ($10^{-2}$)} \\
& \multicolumn{2}{c|}{($10^{-3}$)} & $\mu$ & $\sigma$ & 0.4\% & 0.6\% & 0.8\% & 1.0\% & 1.2\% \\
\hline\hline
PU-Net & 0.83 & 10.58 & 9.52 & 7.98 & 52.38 & 29.59 & 22.51 & 19.86 & 19.08 \\
AR-GCN & 0.42 & 4.75 & 2.73 & 3.22 & 51.20 & 22.17 & \textcolor{blue}{13.89} & 11.18 & 10.57 \\
3PU & 0.44 & 3.04 & \textcolor{red}{0.94} & \textcolor{red}{1.32} & 47.37 & \textcolor{red}{16.44} & \textcolor{red}{8.30} & \textcolor{red}{6.06} & \textcolor{red}{5.89} \\
PU-GAN & 0.52 & 5.66 & 1.45 & 2.61 & 73.55 & 28.63 & 15.35 & \textcolor{blue}{10.69} & \textcolor{blue}{9.20} \\
Dis-PU & 0.47 & 4.44 & \textcolor{blue}{1.45} & 2.47 & 80.53 & 32.32 & 17.69 & 12.36 & 10.52 \\
PU-GCN & 0.69 & 8.65 & 1.89 & 3.03 & 64.55 & 26.22 & 14.94 & 10.98 & 9.73 \\
\hline
ZSPU & \textcolor{blue}{0.33} & \textcolor{blue}{2.40} & 1.56 & 1.62 & \textcolor{red}{38.89} & \textcolor{blue}{19.50} & 14.29 & 12.80 & 12.76 \\
ZSPU-I & \textcolor{red}{0.32} & \textcolor{red}{2.35} & 1.45 & \textcolor{blue}{1.40} & \textcolor{blue}{44.68} & 20.79 & 14.32 & 12.43 & 12.27 \\
\hline
\end{tabular}}
\end{center}
\caption{Quantitative comparisons with supervised upsampling networks on ${Data}_{PS}$.}
\label{tab:quantps}
\end{table}

Table~\ref{tab:quantps} lists the quantitative results with other supervised learning methods on unseen shape categories from ${Data}_{PS}$. Our approach achieves the best global and local results. Globally, the reconstruction is outperformed by our method by a large margin ($21\%$ in CD and HD). Locally, our method achieves the best local uniformity in the smaller disk ($0.4\%$), indicating that ZSPU is compatible to numerous shapes.

\subsection{Qualitative Comparison}
In this section, we qualitatively compare our model with the top performers based on the quantitative results on mesh reconstructions, complex shapes, and real-world scenarios. More results are provided in the supplementary material.

\begin{figure}[h!]
\centering
\begin{subfigure}{\wqreal\textwidth}
\centering
  \makebox[0pt][r]{\makebox[10pt]{\raisebox{40pt}{\rotatebox[origin=c]{90}{Input}}}}%
  \includegraphics[height=\hqreal]{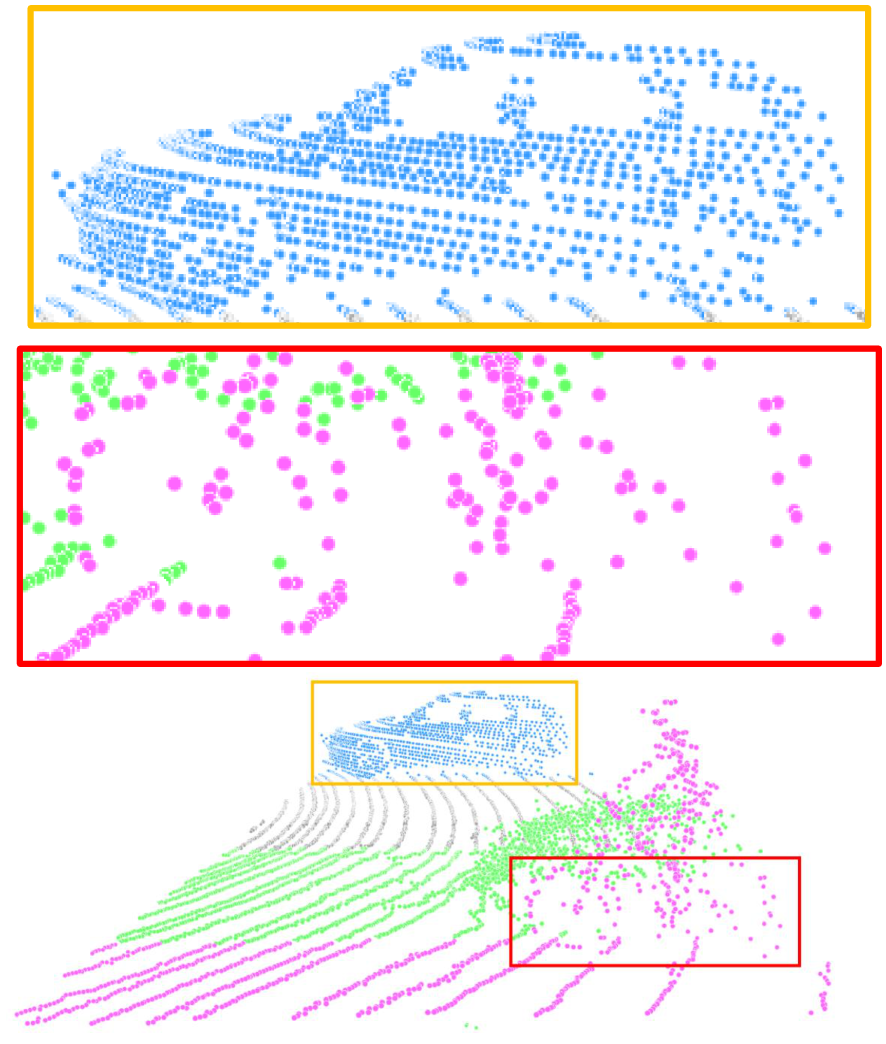}
  \makebox[0pt][r]{\makebox[10pt]{\raisebox{40pt}{\rotatebox[origin=c]{90}{PU-GAN}}}}%
  \includegraphics[height=\hqreal]{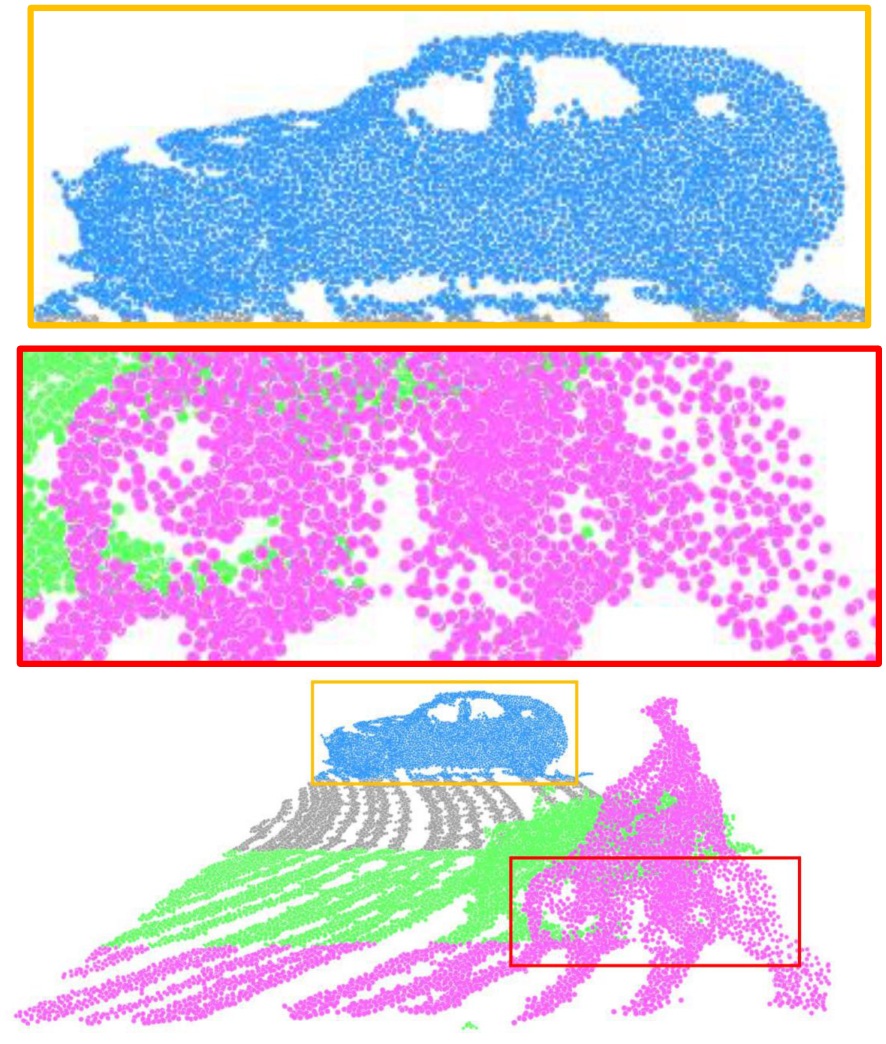}
  \makebox[0pt][r]{\makebox[10pt]{\raisebox{40pt}{\rotatebox[origin=c]{90}{ZSPU}}}}%
  \includegraphics[height=\hqreal]{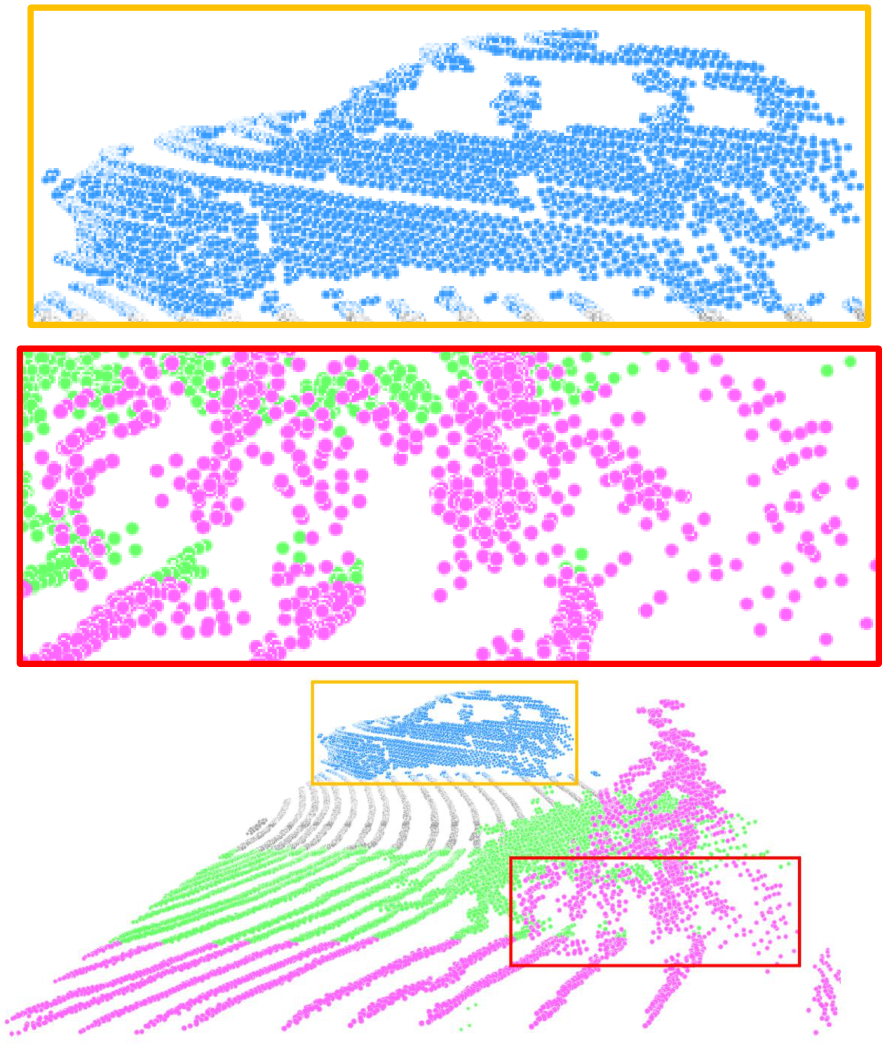}
  \caption{View 1}
  \label{fig3:v1}
\end{subfigure}
\begin{subfigure}{\wqreal\textwidth}
\centering
  \includegraphics[height=\hqreal]{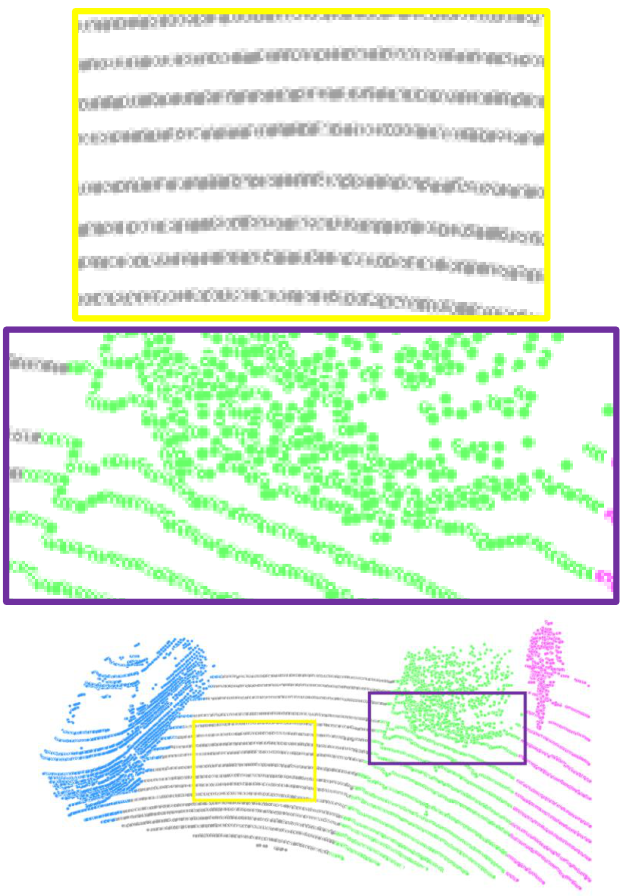}
  \includegraphics[height=\hqreal]{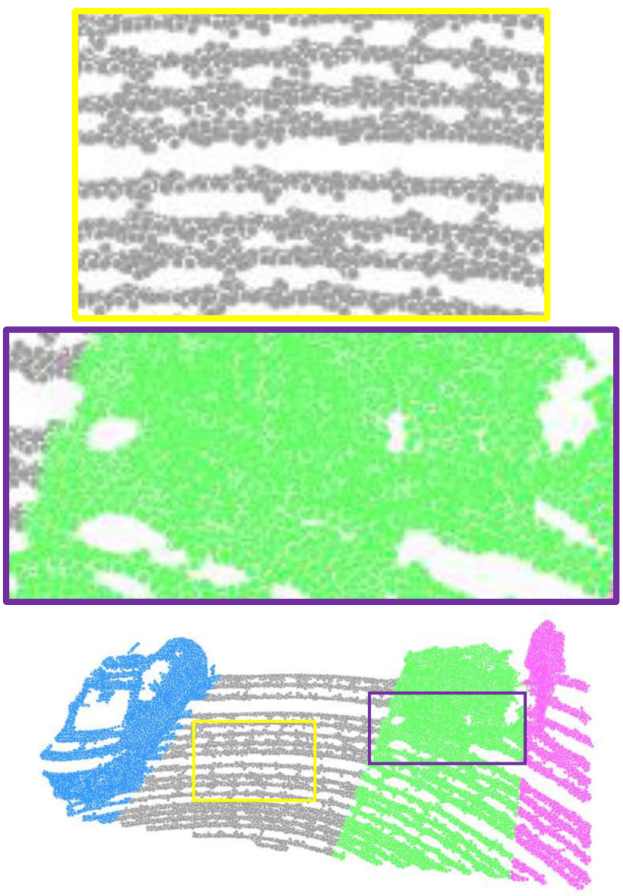}  
  \includegraphics[height=\hqreal]{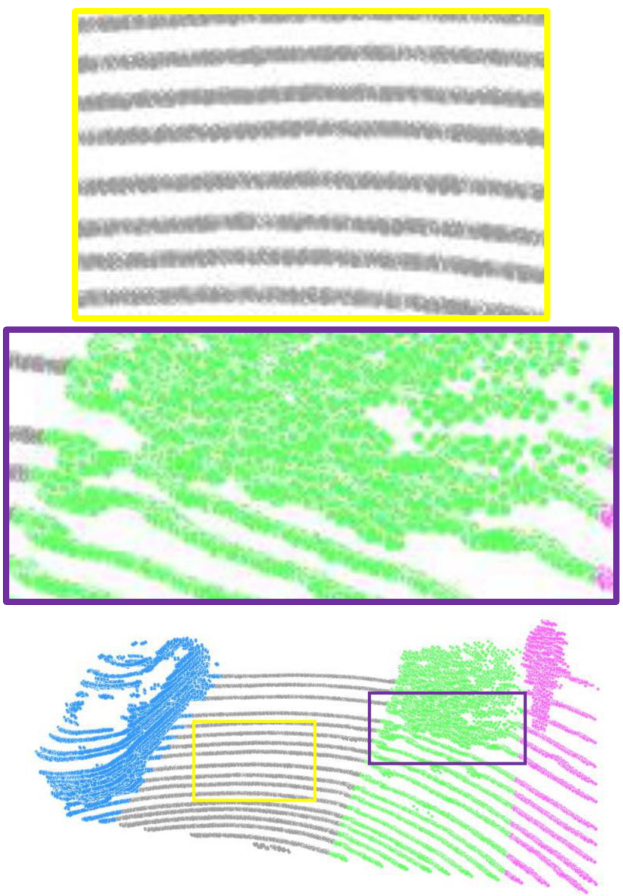}
  \caption{View 2}
  \label{fig3:v2}
\end{subfigure}
\caption{Comparison Between PU-GAN and ZSPU using a real scene containing multiple objects.}
\label{fig:figq3}
\end{figure}

Despite 3PU and PU-GAN achieving best performance regarding uniformity in quantitative results on ${Data}_{PU}$, through visual inspection, we find that they both sacrifice the smoothness of close surfaces in the patch-based prediction phase, which may cause severe mistakes in classification and remeshing tasks. Fig.~\ref{fig:figq} shows that the uniformity metric does not necessarily lead to better surface reconstruction, partly due to the interference between nearby patches. Furthermore, we show some upsampled examples in Fig.~\ref{fig:figq2} for complex shapes in ${Data}_{PS}$. Our model produces significantly better results, particularly in fine detailed areas. These examples demonstrate that ZSPU is more suitable than patch-based upsampling methods on shapes with complex details or rapid changing curvatures.

\textbf{External vs. Internal.}
To further illustrate the upsampling preferences of the external and internal methods on local regions of a point cloud, in Fig.~\ref{fig:errormap} we provide the error map comparison for two examples selected from ${Data}_{PU}$ and ${Data}_{PS}$, respectively. The colors in error maps reveal the nearest distance from each target point to the generated point. Visually, ZSPU generates the least error on local curvatures while maintaining competitive performance on smooth regions.

\textbf{Upsampling Real-Scanned Data.}
PU-GAN achieved the best overall performance on the most common upsampling dataset ${Data}_{PU}$. In Fig.~\ref{fig:figq3}, we further compare our model with PU-GAN on LiDAR-scanned point clouds from the KITTI dataset~\cite{geiger2013vision}, which contains mixed shapes of car, bike, scrub, and ground. NoP is carefully set up for PU-GAN to ensure that all points can be covered. Albeit PU-GAN fills the tiny gaps through the optimization of the uniformity metric, in practice it leads to the potential interference between different parts of an object (\eg, the zoomed frames in Fig.~\ref{fig:figq3}). These examples clearly demonstrate the advantages of our ``zero-shot'' model over supervised patch-based methods in real-world scenarios.

\section{Conclusion}
In this paper, we present a ``zero-shot'' point cloud upsampling framework, which takes the entire point cloud as input and trains the network internally to produce a point cloud with a higher resolution. Through the holistic and internal learning design, our model avoids the intricate settings used in patch-based approaches and achieves competitive/superior results on benchmark datasets. It is well suited to handle high curvature regions or complex scenes obtained from real-world scans.

\noindent \textbf{Acknowledgment}: This research was partly supported by the National Science Foundation (NSF: \#1948338) and the Department of Defense (DoD: \#W81XWH-21-1-0570).

\bibliographystyle{IEEEbib}
\bibliography{icme2022}

\begin{thebibliography}{10}

\bibitem{yu2018pu}
Lequan Yu, Xianzhi Li, Chi-Wing Fu, Daniel Cohen-Or, and Pheng-Ann Heng,
\newblock ``{PU-Net}: Point cloud upsampling network,''
\newblock in {\em Proc. of the IEEE/CVF Conf. on Computer Vision and Pattern
  Recognition (CVPR)}, 2018, pp. 2790--2799.

\bibitem{yu2018ec}
Lequan Yu, Xianzhi Li, Chi-Wing Fu, Daniel Cohen-Or, and Pheng-Ann Heng,
\newblock ``{EC-Net}: an edge-aware point set consolidation network,''
\newblock in {\em Proc. of the European Conf. on Computer Vision (ECCV)}, 2018,
  pp. 386--402.

\bibitem{yifan2019patch}
Wang Yifan, Shihao Wu, Hui Huang, Daniel Cohen-Or, and Olga Sorkine-Hornung,
\newblock ``Patch-based progressive 3d point set upsampling,''
\newblock in {\em Proc. of the IEEE/CVF Conf. on Computer Vision and Pattern
  Recognition (CVPR)}, 2019, pp. 5958--5967.

\bibitem{li2019pugan}
Ruihui Li, Xianzhi Li, Chi-Wing Fu, Daniel Cohen-Or, and Pheng-Ann Heng,
\newblock ``{PU-GAN}: a point cloud upsampling adversarial network,''
\newblock in {\em Proc. of the IEEE/CVF Int. Conf. on Computer Vision (ICCV)},
  2019, pp. 7203--7212.

\bibitem{li2021point}
Ruihui Li, Xianzhi Li, Pheng-Ann Heng, and Chi-Wing Fu,
\newblock ``Point cloud upsampling via disentangled refinement,''
\newblock in {\em Proc. of the IEEE/CVF Conf. on Computer Vision and Pattern
  Recognition (CVPR)}, 2021, pp. 344--353.

\bibitem{qian2021pu}
Guocheng Qian, Abdulellah Abualshour, Guohao Li, Ali Thabet, and Bernard
  Ghanem,
\newblock ``{PU-GCN}: Point cloud upsampling using graph convolutional
  networks,''
\newblock in {\em Proc. of the IEEE/CVF Conf. on Computer Vision and Pattern
  Recognition (CVPR)}, 2021, pp. 11683--11692.

\bibitem{he2019geonet}
Tong He, Haibin Huang, Li~Yi, Yuqian Zhou, Chihao Wu, Jue Wang, and Stefano
  Soatto,
\newblock ``{GeoNet}: Deep geodesic networks for point cloud analysis,''
\newblock in {\em Proc. of the IEEE/CVF Conf. on Computer Vision and Pattern
  Recognition (CVPR)}, 2019, pp. 6888--6897.

\bibitem{qi2017pointnet}
Charles~R Qi, Hao Su, Kaichun Mo, and Leonidas~J Guibas,
\newblock ``{PointNet}: Deep learning on point sets for 3d classification and
  segmentation,''
\newblock in {\em Proc. of the IEEE Conf. on Computer Vision and Pattern
  Recognition (CVPR)}, 2017, pp. 652--660.

\bibitem{qi2017pointnet++}
Charles~Ruizhongtai Qi, Li~Yi, Hao Su, and Leonidas~J Guibas,
\newblock ``{PointNet++}: Deep hierarchical feature learning on point sets in a
  metric space,''
\newblock {\em Advances in Neural Information Processing Systems}, vol. 30,
  2017.

\bibitem{wu2019point}
Huikai Wu, Junge Zhang, and Kaiqi Huang,
\newblock ``Point cloud super resolution with adversarial residual graph
  networks,''
\newblock in {\em British Machine Vision Conf. (BMVC)}, 2020.

\bibitem{ma2020structure}
Cheng Ma, Yongming Rao, Yean Cheng, Ce~Chen, Jiwen Lu, and Jie Zhou,
\newblock ``Structure-preserving super resolution with gradient guidance,''
\newblock in {\em Proc. of the IEEE/CVF Conf. on Computer Vision and Pattern
  Recognition (CVPR)}, 2020, pp. 7769--7778.

\bibitem{niu2020single}
Ben Niu, Weilei Wen, Wenqi Ren, Xiangde Zhang, Lianping Yang, Shuzhen Wang,
  Kaihao Zhang, Xiaochun Cao, and Haifeng Shen,
\newblock ``Single image super-resolution via a holistic attention network,''
\newblock in {\em Proc. of the European Conf. on Computer Vision (ECCV)}, 2020,
  pp. 191--207.

\bibitem{zhang2019zoom}
Xuaner Zhang, Qifeng Chen, Ren Ng, and Vladlen Koltun,
\newblock ``Zoom to learn, learn to zoom,''
\newblock in {\em Proc. of the IEEE/CVF Conf. on Computer Vision and Pattern
  Recognition (CVPR)}, 2019, pp. 3762--3770.

\bibitem{zhang2019image}
Zhifei Zhang, Zhaowen Wang, Zhe Lin, and Hairong Qi,
\newblock ``Image super-resolution by neural texture transfer,''
\newblock in {\em Proc. of the IEEE/CVF Conf. on Computer Vision and Pattern
  Recognition (CVPR)}, 2019, pp. 7982--7991.

\bibitem{shocher2018zero}
Assaf Shocher, Nadav Cohen, and Michal Irani,
\newblock ````{Zero-Shot}'' super-resolution using deep internal learning,''
\newblock in {\em Proc. of the IEEE/CVF Conf. on Computer Vision and Pattern
  Recognition (CVPR)}, 2018, pp. 3118--3126.

\bibitem{shilane2004princeton}
Philip Shilane, Patrick Min, Michael Kazhdan, and Thomas Funkhouser,
\newblock ``The {Princeton} shape benchmark,''
\newblock in {\em Proc. of Shape Modeling Applications}, 2004, pp. 167--178.

\bibitem{geiger2013vision}
Andreas Geiger, Philip Lenz, Christoph Stiller, and Raquel Urtasun,
\newblock ``Vision meets robotics: The {KITTI} dataset,''
\newblock {\em The Int. Journal of Robotics Research}, vol. 32, no. 11, pp.
  1231--1237, 2013.

\end{thebibliography}

\newpage
\appendix

\section{Additional Failure Cases of Patch-based Methods}
\begin{figure*}[h!]
\centering
\captionsetup[subfigure]{labelformat=empty}
\begin{subfigure}{\wfail\textwidth}
\centering
  \makebox[0pt][r]{\makebox[20pt]{\raisebox{40pt}{\rotatebox[origin=c]{90}{$r_p=1$}}}}%
  \includegraphics[height=\hefail]{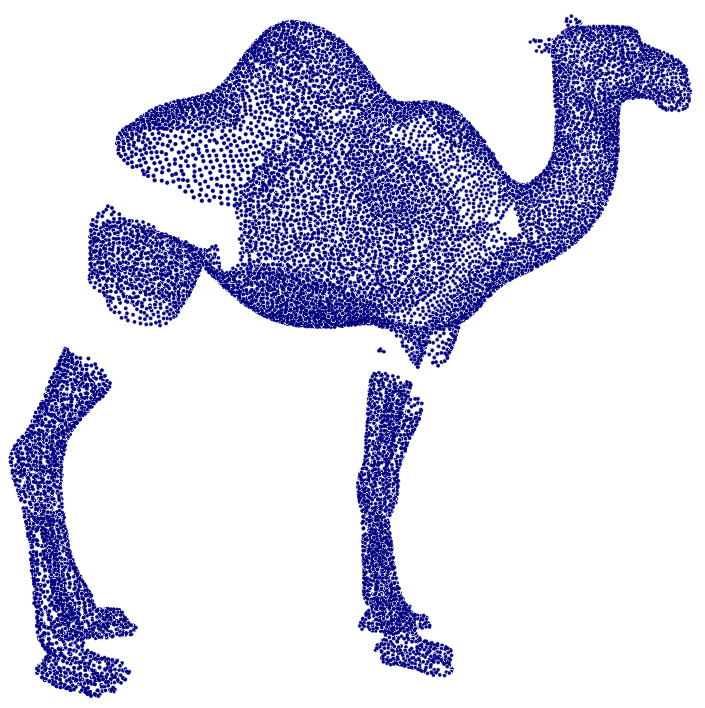}
  \makebox[0pt][r]{\makebox[20pt]{\raisebox{40pt}{\rotatebox[origin=c]{90}{$r_p=2$}}}}%
  \includegraphics[height=\hefail]{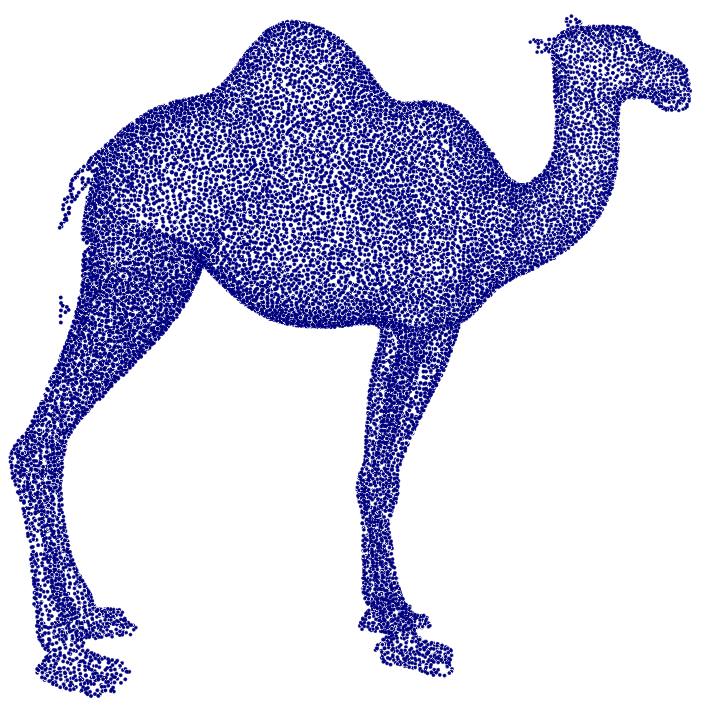}
  \makebox[0pt][r]{\makebox[20pt]{\raisebox{50pt}{\rotatebox[origin=c]{90}{$r_p=1$}}}}%
  \includegraphics[height=\hefailb]{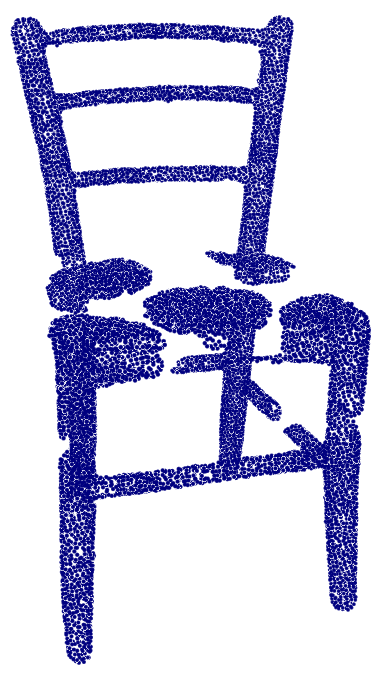}
  \makebox[0pt][r]{\makebox[20pt]{\raisebox{50pt}{\rotatebox[origin=c]{90}{$r_p=2$}}}}%
  \includegraphics[height=\hefailb]{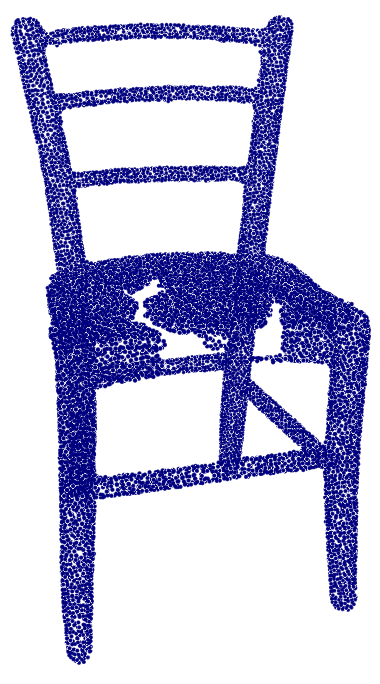}
  \caption{3PU}
\end{subfigure}
\begin{subfigure}{\wfail\textwidth}
\centering
  \includegraphics[height=\hefail]{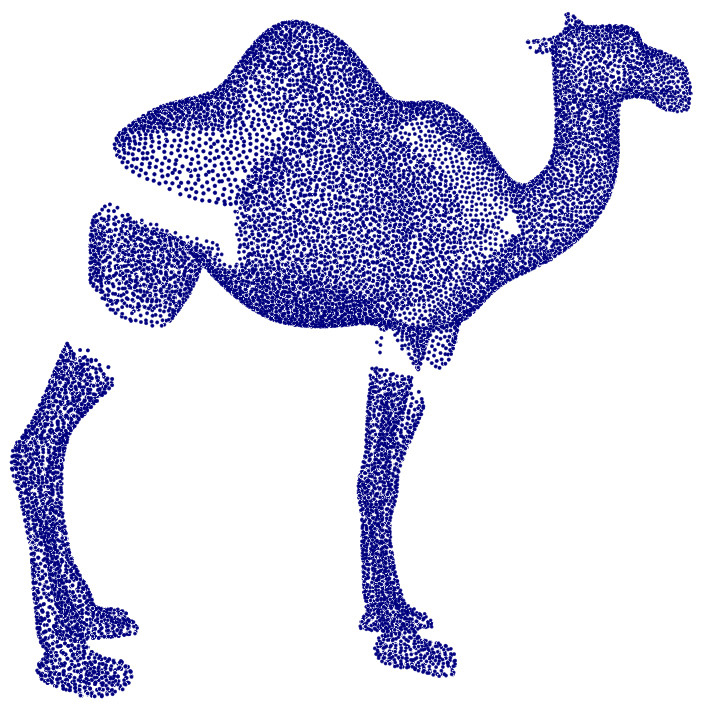}
  \includegraphics[height=\hefail]{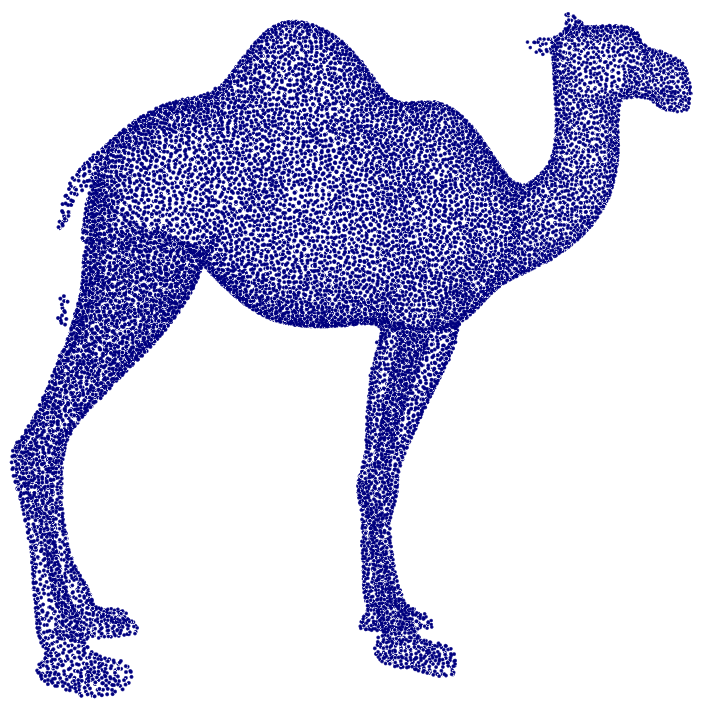}  
  \includegraphics[height=\hefailb]{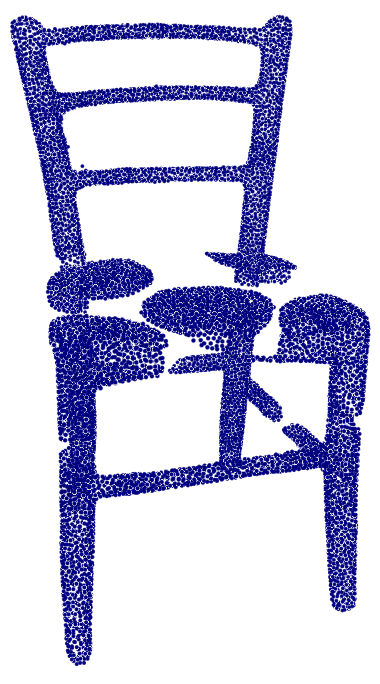}
  \includegraphics[height=\hefailb]{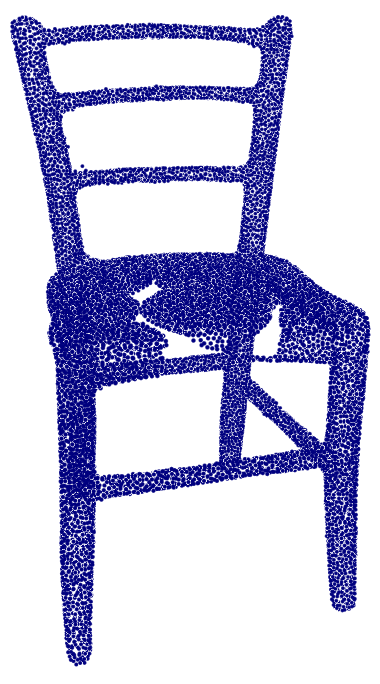}
  \caption{Dis-PU}
\end{subfigure}
\begin{subfigure}{\wfail\textwidth}
\centering
  \includegraphics[height=\hefail]{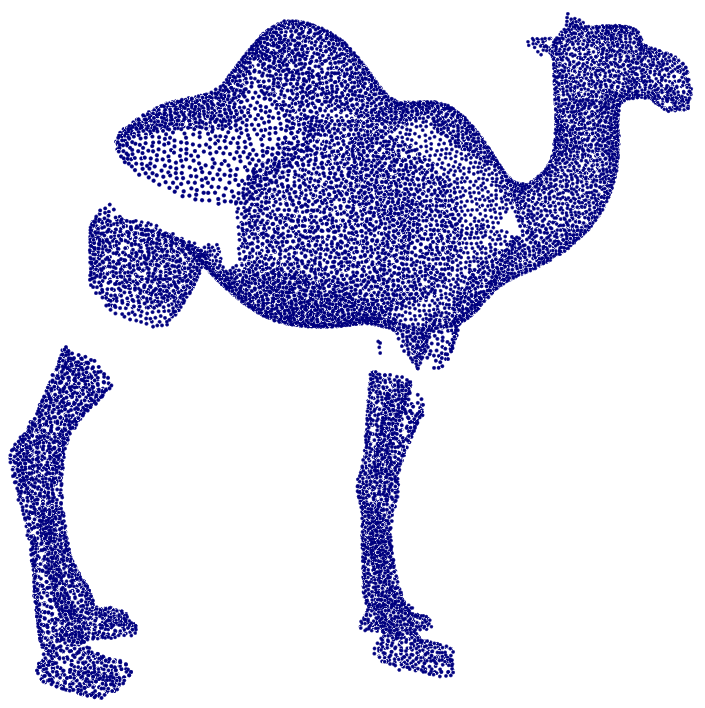}
  \includegraphics[height=\hefail]{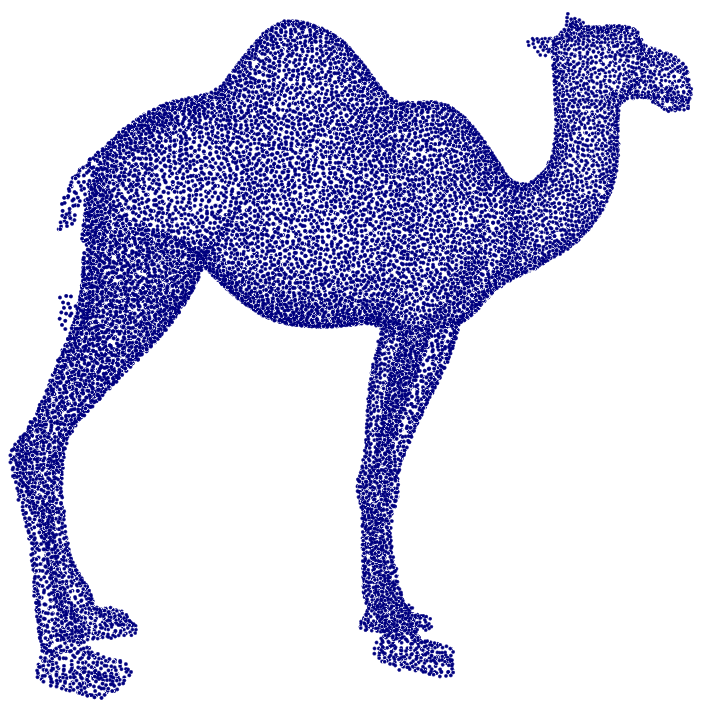}  
  \includegraphics[height=\hefailb]{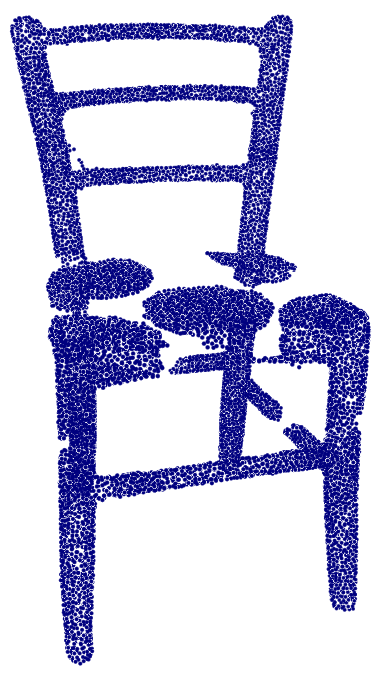}
  \includegraphics[height=\hefailb]{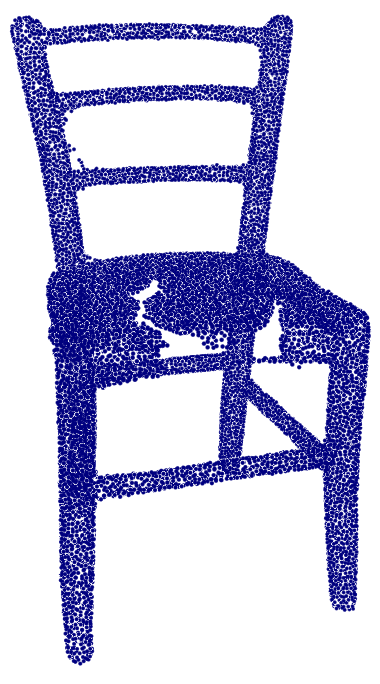}
  \caption{PU-GAN}
\end{subfigure}
\begin{subfigure}{\wfail\textwidth}
\centering
  \includegraphics[height=\hefail]{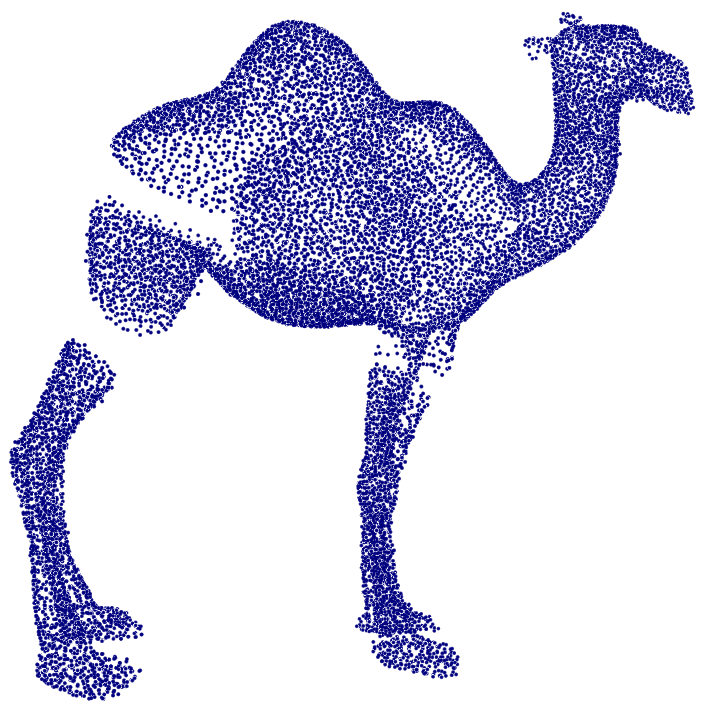}
  \includegraphics[height=\hefail]{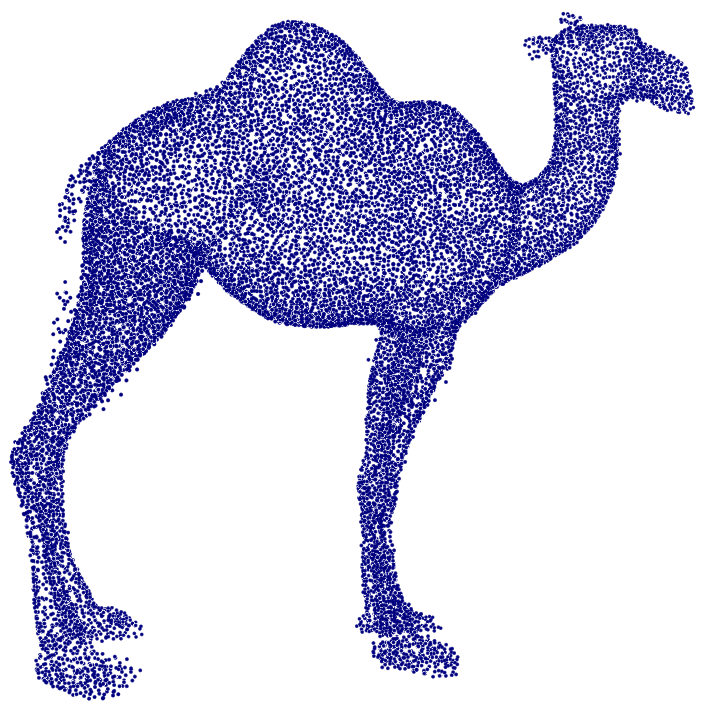}  
  \includegraphics[height=\hefailb]{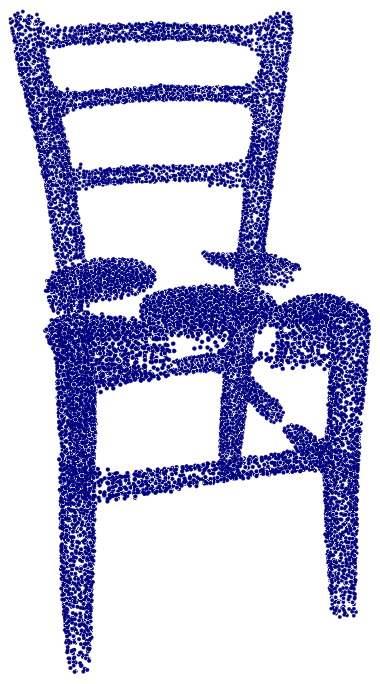}
  \includegraphics[height=\hefailb]{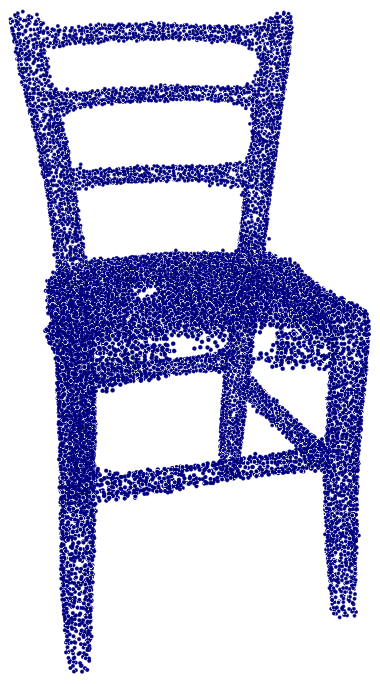}
  \caption{PU-GCN}
\end{subfigure}
\caption{Failure cases of patch-based methods.}
\label{fig:patchfail}
\end{figure*}

The centroids found by farthest-point-sampling (FPS) algorithm to generate patches are not always enough to cover the entire shape of point clouds. We show the failure cases of patch-based methods, \ie 3PU, Dis-PU, PU-GAN and PU-GCN, caused by not enough number of patches (NoP) in Fig.~\ref{fig:patchfail}, where $r_p$ is the ratio of patch numbers and will be multiplied by NoP.

\section{Generator}
\begin{figure}[h!]
\begin{center}
    \includegraphics[width=0.9\linewidth]{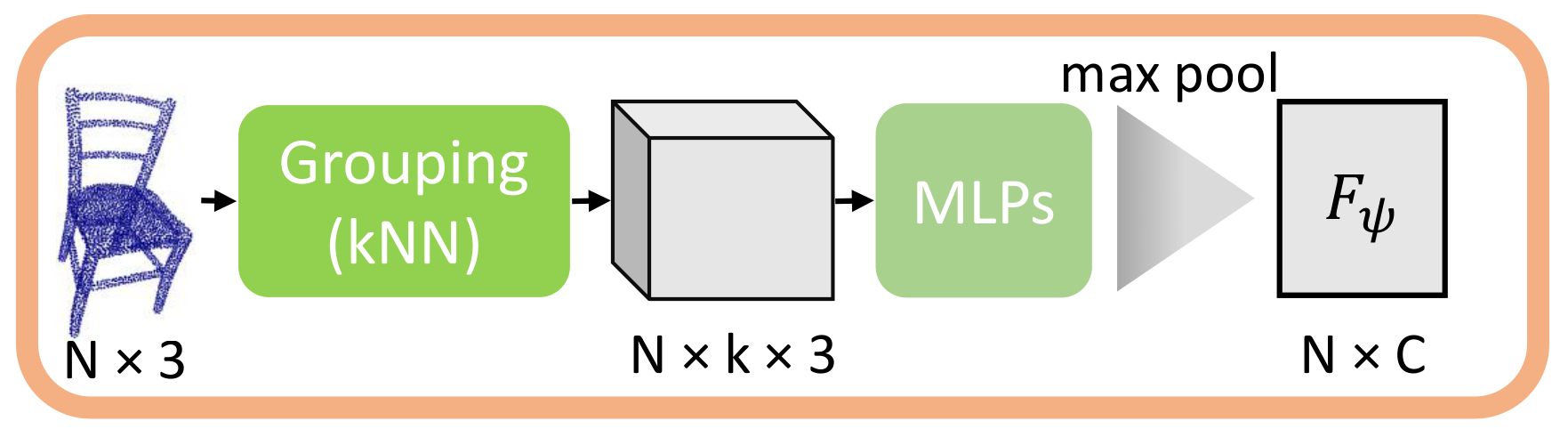}
\end{center}
    \caption{The graph feature extraction component.}
\label{fig:gmlp}
\end{figure}

\subsection{Graph feature extraction}
Empirically, using the coordinate information and its surrounding features can produce well-shaped point clouds, especially for those with crowd surfaces in a certain region. For this reason, we adopt the component in AR-GCN that learns the convolution-like relationships between each point and its k nearest neighbors (kNN). As described by Fig.~\ref{fig:gmlp}, we group the $N\times k \times 3$ graphical information by kNN, which further forms the $N \times C$ feature $F_\psi=\{\sum_{k=1}^{K} \psi(x_k) / K|x_k \in H(x_i), x_i \in \chi \}$ via multi-layer perceptrons (MLPs) averaged by closest features of each point, where $H(x_i)$ denotes the neighbor set of point $x_i$ and $\psi(x_k)$ is the latent feature of $x_k$.

\begin{figure}[h!]
\begin{center}
    \includegraphics[width=0.9\linewidth]{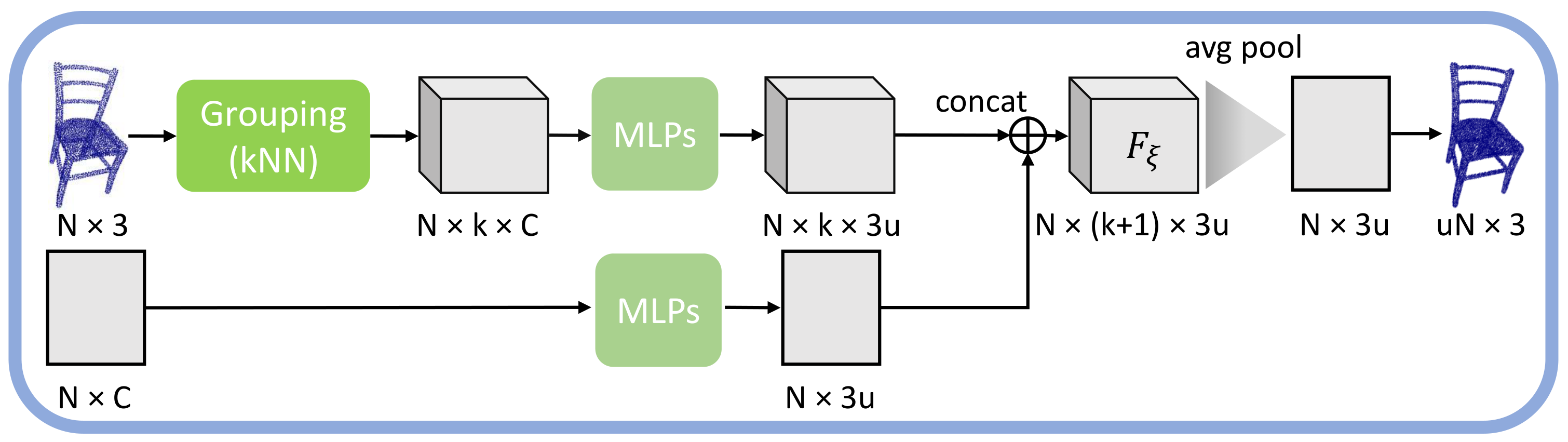}
\end{center}
    \caption{The up expansion component.}
\label{fig:up}
\end{figure}

\subsection{Up expansion}
This component is designed to be flexible to varying point size, such that the point clouds and latent representations can be successively upsampled with a small ratio $u$. As illustrated in Fig.~\ref{fig:up}, the point set $\chi$ and its graphical feature are fed to two parallel layers respectively, producing one tensor of dimension $(N \times k \times 3u)$ and another of dimension $(N \times 3u)$. Then we average concatenated feature $F_\xi$ based on the center and neighbor embedding. $\chi'$ is obtained after reshaping $F_\xi$. The generated $\chi'$ is finally fed to a graph feature extraction component to exploit neighbor features for subsequent up expansions.


\section{Detail of Loss Functions}
\subsection{Reconstruction loss}
Earth mover distance (EMD) is a metric between the generated point set $\chi^r$ and  ground truth $\chi^{gt}$ based on the minimal cost of transiting points from $\chi^r \in \mathbb{R}^3$ to $\chi^{gt} \in \mathbb{R}^3$:
\begin{equation}
L_{EMD} = \underset{\phi : \chi^r \rightarrow \chi^{gt}}{min} \sum_{x \in \chi^r} {||x-\phi(x)||},
\end{equation}
where $\phi : \chi^r \rightarrow \chi^{gt}$ indicates the bijection mapping, and $||\cdot||$ represents the L2-norm. Note that the terminology of ground truth here also represents the input because of internal learning. 

\subsection{Repulsion loss}
Repulsion loss aims to push the points far away from each other if a local region (a query ball or kNN) contains too many of them:
\begin{equation}
L_{rep}(\chi^r) = \frac{1}{N^r \cdot K} \sum_{i=1}^{N^r} \sum_{k=1}^{K} \eta (||x_i - x_{ik}||),
\end{equation}
in which, $K$ is the number of neighborhoods we choose, $\eta (r) = max(0, h-r)$ is a function to penalize point $x_i$ if its neighbors are too close to it, and h is the empirical threshold that controls the penalization's effect only when the distance of two points is too close. In general,
$L_{rep}$ tries to distribute the points uniformly by penalizing points that are too close to each other.

\subsection{Uniform loss}
To uniformly upsample points, averaging the number of points in small disks alleviates the non-uniform distribution, and the expected number of points can be approximated from the proportion between small disks and the overall area. However, this is not suitable at holistic scale, since the shape in each disk might vary rapidly. Besides, for the purpose of enhancing training efficiency, we only apply the approximate point-to-neighbor distance as the metric in our uniform loss. Specifically, $M$ seed points are selected by farthest point sampling (FPS) on $\chi^r$, forming $M$ disks ($\chi_j \in \chi^r, j=1,2,...,M$) with a small radius $r_d$, where $r_d=\sqrt{p \cdot A}$ for each $p \in \{0.004, 0.006, 0.008, 0.01, 0.012\}$ in a unit area $A$. In each disk, we use an approximate square to estimate distances in each region, such that the distance between each point is expected to be $\sqrt{\pi r_d^2 / K}$ if the region is small enough. Thus, the uniform loss is formulated as:
\begin{equation}\label{eq:uniform}
L_{uni} = \sum_{j=1}^{M} \sum_{k=1}^{K} \frac{||d_{jk}-\sqrt{\pi r_d^2 / K}||^2}{\sqrt{\pi r_d^2 / K}},
\end{equation}
where $d_{jk}$ is the distance between point $j$ with the $k$th neighbor. 

\subsection{Adversarial loss}
We adopt the discriminator structure from PU-GAN, where the loss functions of the generative network $G$ and the discriminative network $D$ can formally be written as:
\begin{equation}
L_G = ||1-{log}D(\chi^r)||^2,
\end{equation}
\begin{equation}
L_D = ||{log}D(\chi^r)||^2 + ||1-{log}D (\chi^{gt})||^2.
\end{equation}

\section{Additional Qualitative Results}

\subsection{Upsampling with varying input sizes}
\begin{figure}[h!]
\centering
\captionsetup[subfigure]{labelformat=empty}
\begin{subfigure}{\wden\textwidth}
\centering
  \caption{4096 points}
  \makebox[0pt][r]{\makebox[20pt]{\raisebox{20pt}{\rotatebox[origin=c]{90}{Input}}}}%
  \includegraphics[height=\heden]{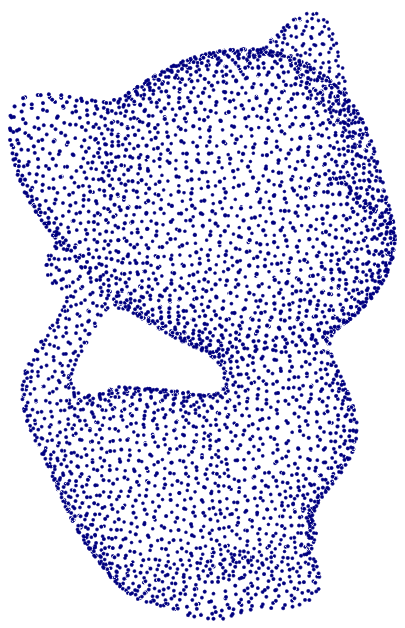}
  \makebox[0pt][r]{\makebox[20pt]{\raisebox{20pt}{\rotatebox[origin=c]{90}{Non-Ideal}}}}%
  \includegraphics[height=\heden]{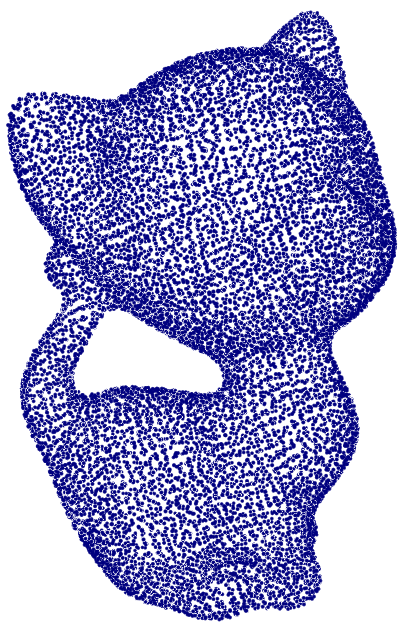}
  \makebox[0pt][r]{\makebox[20pt]{\raisebox{20pt}{\rotatebox[origin=c]{90}{Ideal}}}}%
  \includegraphics[height=\heden]{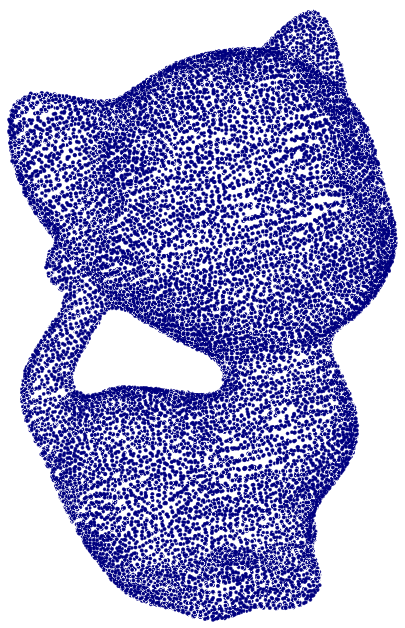}
  \caption{96s}
\end{subfigure}
\begin{subfigure}{\wden\textwidth}
\centering
  \caption{2048 points}
  \includegraphics[height=\heden]{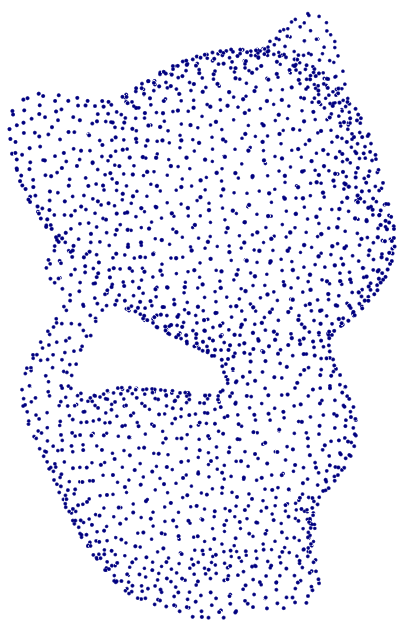}
  \includegraphics[height=\heden]{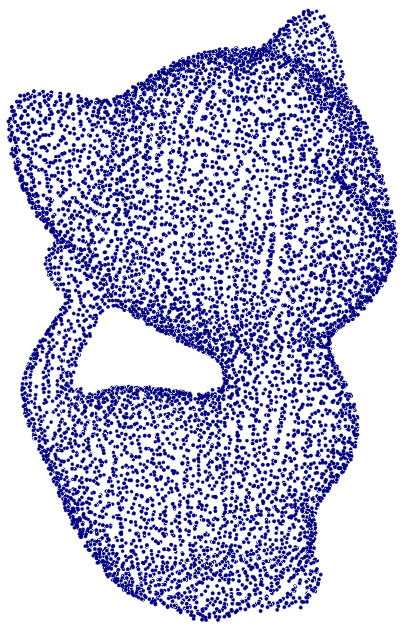}  
  \includegraphics[height=\heden]{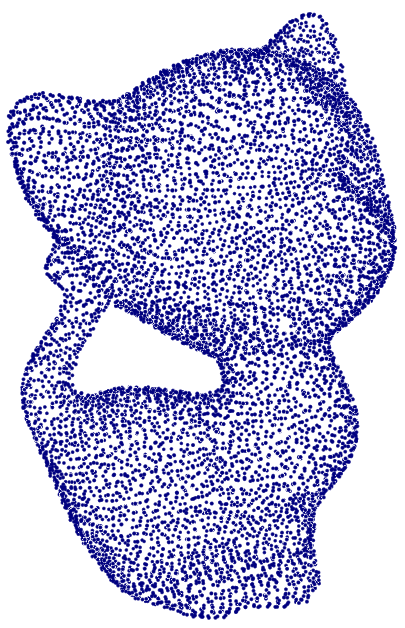}
  \caption{43s}
\end{subfigure}
\begin{subfigure}{\wden\textwidth}
\centering
  \caption{1024 points}
  \includegraphics[height=\heden]{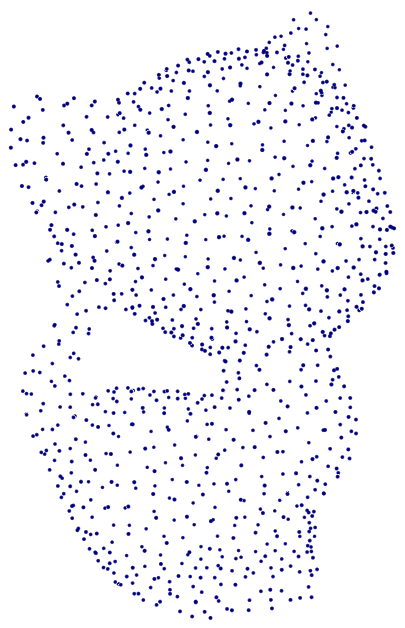}
  \includegraphics[height=\heden]{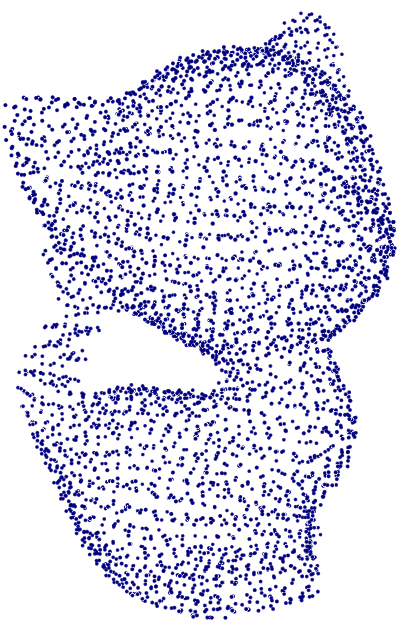}  
  \includegraphics[height=\heden]{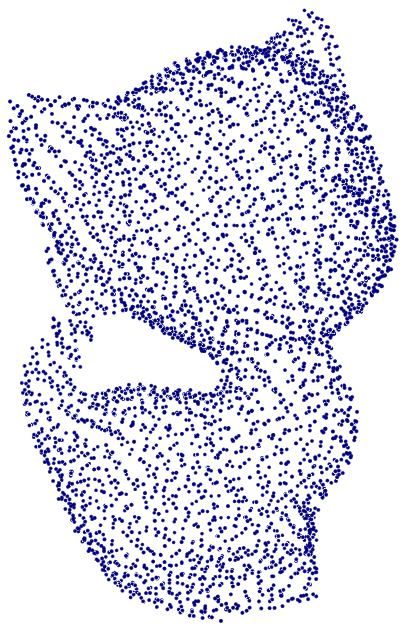}
  \caption{26s}
\end{subfigure}
\begin{subfigure}{\wden\textwidth}
\centering
  \caption{512 points}
  \includegraphics[height=\heden]{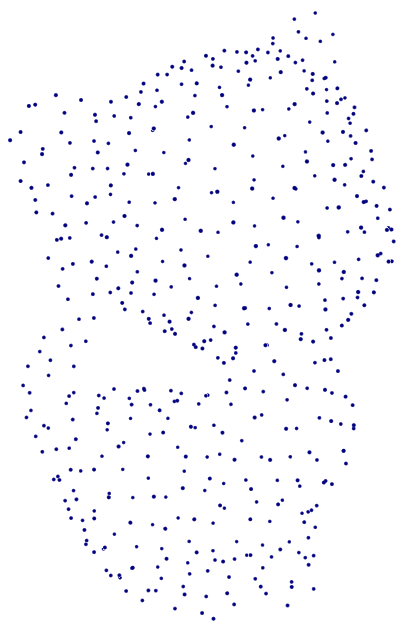}
  \includegraphics[height=\heden]{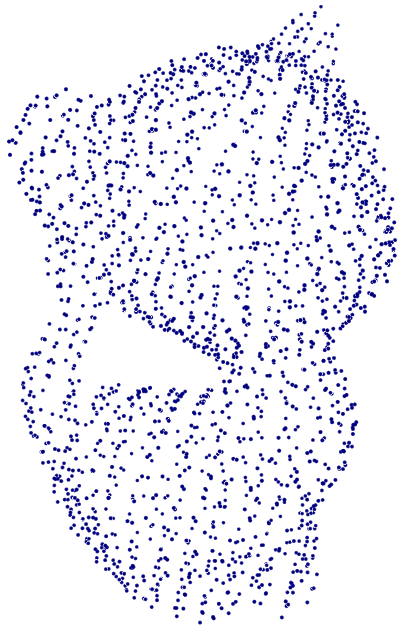}  
  \includegraphics[height=\heden]{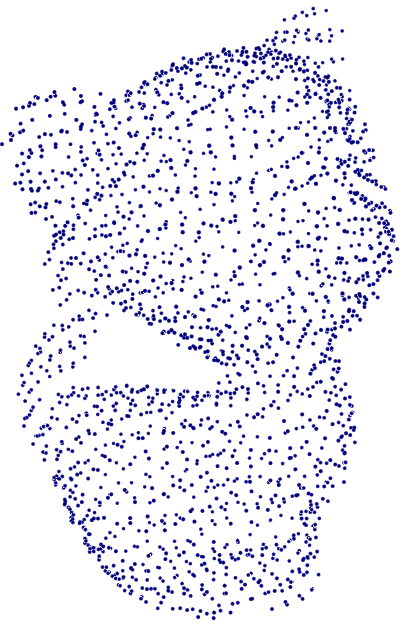}
  \caption{18s} 
\end{subfigure}
\caption{Results for varying densities.}
\label{fig:density}
\end{figure}
\begin{figure}[h!]
\centering
\captionsetup[subfigure]{labelformat=empty}
\begin{subfigure}{\wnoi\textwidth}
\centering
  \makebox[0pt][r]{\makebox[20pt]{\raisebox{20pt}{\rotatebox[origin=c]{90}{Input}}}}%
  \includegraphics[height=\henoi]{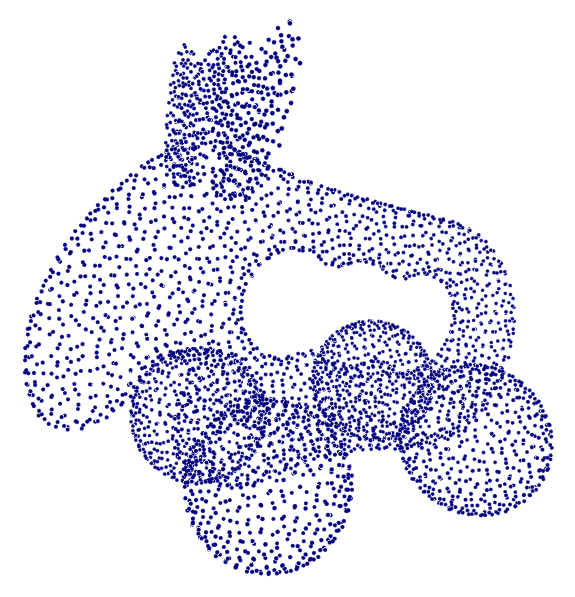}
  \makebox[0pt][r]{\makebox[20pt]{\raisebox{20pt}{\rotatebox[origin=c]{90}{Non-Ideal}}}}%
  \includegraphics[height=\henoi]{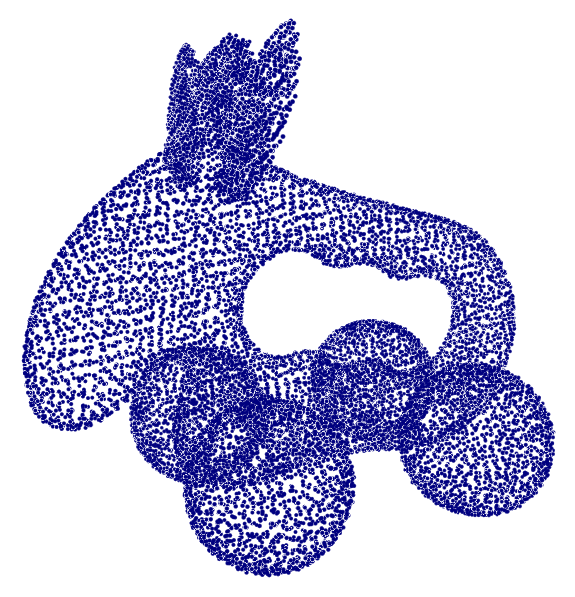}
  \makebox[0pt][r]{\makebox[20pt]{\raisebox{20pt}{\rotatebox[origin=c]{90}{Ideal}}}}%
  \includegraphics[height=\henoi]{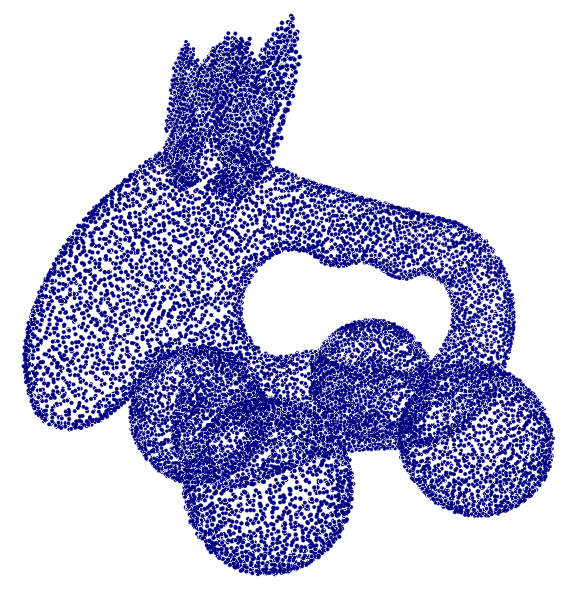}
\end{subfigure}
\begin{subfigure}{\wnoi\textwidth}
\centering
  \includegraphics[height=\henoi]{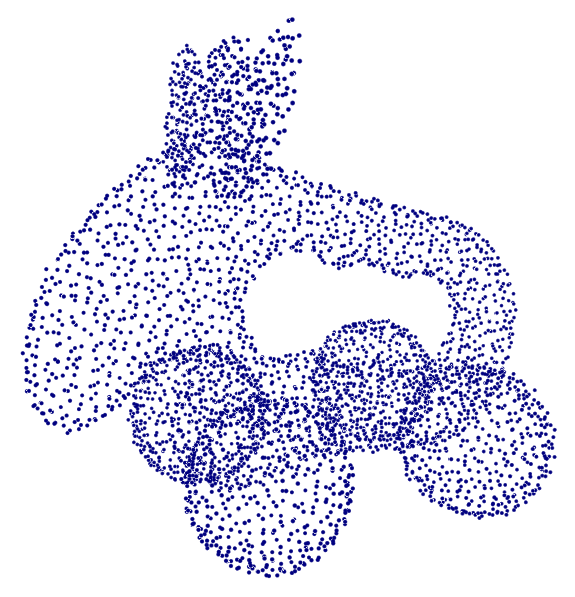}
  \includegraphics[height=\henoi]{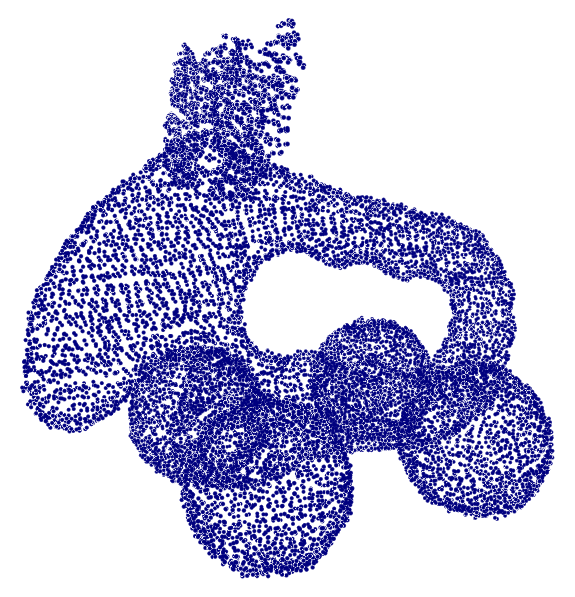}  
  \includegraphics[height=\henoi]{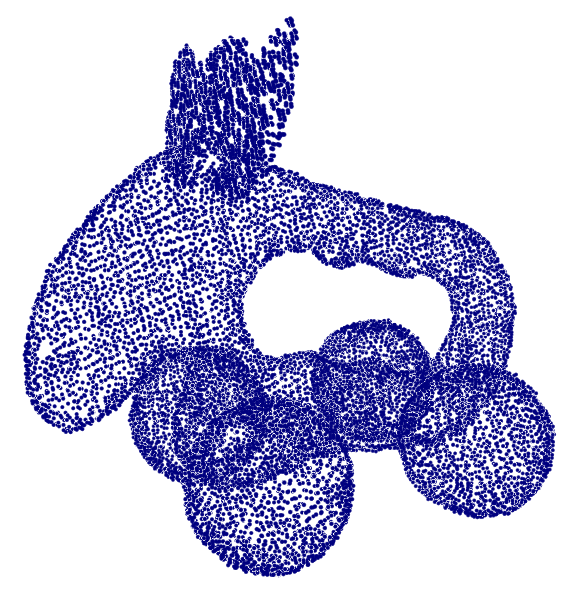}
\end{subfigure}
\begin{subfigure}{\wnoi\textwidth}
\centering
  \includegraphics[height=\henoi]{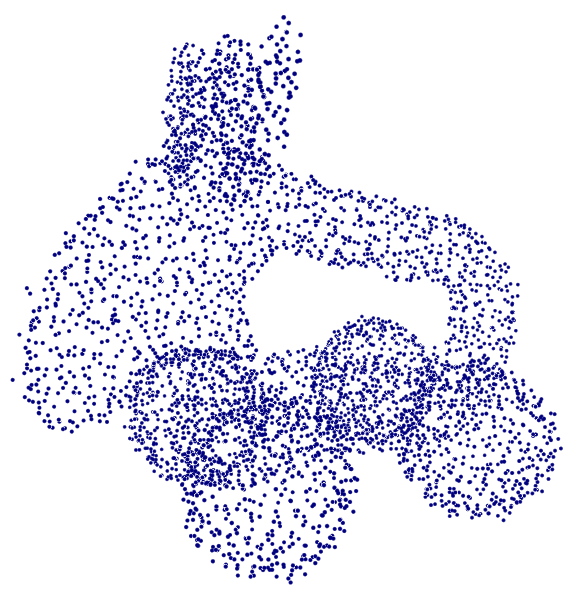}
  \includegraphics[height=\henoi]{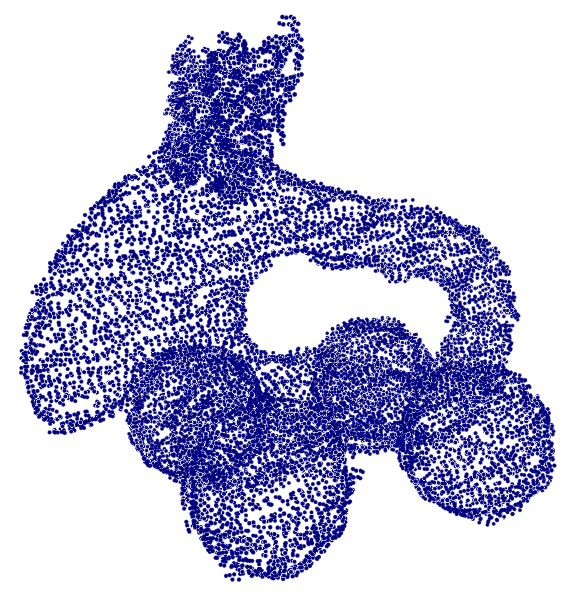}  
  \includegraphics[height=\henoi]{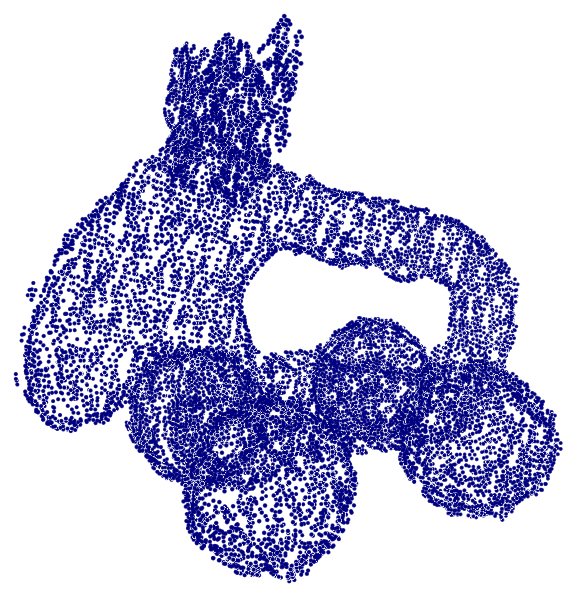}  
\end{subfigure}
\caption{Results for noise levels at $0\%$, $0.5\%$, and $1\%$.}
\label{fig:noise}
\end{figure}

Fig.~\ref{fig:density} shows the results for applying ZSPU on a kitten point cloud with different sizes and the corresponding time cost. As the size becomes smaller, the learning time drops tremendously. Through our visual observation, the ideal kernel performs better than the non-ideal one for larger input sizes, \ie, 2048 and 4096 points.

\subsection{Upsampling with varying noise levels}
Fig.~\ref{fig:noise} shows the results when applying different noise levels, \ie, $0\%$, $0.5\%$, and $1\%$, to the input. When the noise becomes larger, \ie, $1\%$, the result using the non-ideal kernel has visually smoother surface than the one using the ideal kernel. This is mainly because that the downsampled $\chi^{LR}$ using non-ideal kernel itself introduces a small noise level to the training.

\section{Ablation Study}
We conduct the ablation study on ${Data}_{PU}$. In Table~\ref{ab:compo}, by removing the following components respectively: the discriminator (D), the self-attention (Self-Att) module in the discriminator, the progressive manner (PM) for up extensions, the uniform loss ($L_{uni}$), and the repulsion loss ($L_{rep}$), it is clear that all the components and losses are necessary for the proposed method. The result of full version with CD loss (Full-CD) is also provided in Table~\ref{ab:compo}, based on which we selected EMD as our final choice of the reconstruction loss.
The ablation study for various pair sizes is reported in Table~\ref{ab:pair}, and 12 was selected accordingly.

\begin{table}[h!]
\begin{center}
\scalebox{\tablesize}{
 	\begin{tabular}{c || c c c c} 
 	\hline
 	w/o & CD ($10^{-4}$) & HD ($10^{-4}$) & P2F ($10^{-4}$) & Uni ($10^{-4}$)\\
 	\hline\hline
 	D & 2.14 & \textcolor{blue}{11.37} & 23.31 & 21.48 \\ 
 	Self-Att & 2.15 & 11.77 & 22.71 & 21.24 \\
 	PM & 2.15 & 12.23 & 26.96 & 21.81 \\
 	$L_{uni}$ & 2.21 & 12.31 & \textcolor{blue}{22.51} & 22.01 \\
 	$L_{rep}$ & \textcolor{blue}{2.01} & 11.58 & 23.25 & \textcolor{red}{20.42} \\
 	\hline
 	\textbf{Full-CD} & 3.04 & 19.31 & 35.90 & \textcolor{blue}{20.75} \\
 	\textbf{Full} & \textcolor{red}{1.93} & \textcolor{red}{11.09} & \textcolor{red}{21.16} & 22.43 \\
 	\hline
	\end{tabular}}
	\end{center}
\caption{Results without specific components or losses.}
\label{ab:compo}
\end{table}
\begin{table}[h]
\begin{center}
\scalebox{\tablesize}{
 	\begin{tabular}{c || c c c c | c } 
 	\hline
 	\# of Pair & CD ($10^{-4}$) & HD ($10^{-4}$) & P2F ($10^{-4}$) & Uni ($10^{-4}$) & time\\
 	\hline\hline
 	2 & 2.25/1.98 & 13.54/11.13 & 26.44/21.27 & 22.44/15.91 & 60 \\ 
 	6 & 2.26/\textcolor{blue}{1.89} & 12.22/\textcolor{blue}{10.76} & 24.70/19.35 & 20.98/16.18 & 75 \\
 	24 & \textcolor{blue}{2.13}/1.97 & \textcolor{blue}{12.21}/\textcolor{red}{10.74} & 23.70/\textcolor{red}{18.94} & \textcolor{blue}{20.18}/16.22 & 145 \\
 	32 & 2.16/\textcolor{red}{1.88} & 12.26/11.41 & \textcolor{blue}{23.38}/\textcolor{blue}{19.06} & \textcolor{red}{20.04}/\textcolor{red}{15.81} & 180 \\
 	\hline
 	\textbf{12} & \textcolor{red}{1.93}/1.91 & \textcolor{red}{11.09}/11.10 & \textcolor{red}{21.16}/19.01 & 22.43/\textcolor{blue}{15.82} & 100\\
 	\hline
	\end{tabular}}
	\end{center}
    \caption{Results under non-ideal/ideal kernels with various pair sizes.}
	\label{ab:pair}
\end{table}

\end{document}